\documentclass{article}

\PassOptionsToPackage{numbers, compress}{natbib}
\usepackage[preprint]{neurips_2026}

\usepackage[utf8]{inputenc}
\usepackage[T1]{fontenc}
\usepackage{hyperref}
\usepackage{url}
\usepackage{booktabs}
\usepackage{amsfonts}
\usepackage{nicefrac}
\usepackage{microtype}
\usepackage{xcolor}
\usepackage{graphicx}
\usepackage{amsmath}
\usepackage{algorithm}
\usepackage{algorithmic}
\usepackage{multirow}
\usepackage{amssymb}
\usepackage{natbib}
\usepackage{xurl}

\title{Human-inspired, Task-Dimension-Guided Exploration for Efficient Learning in High Dimensions}

\author{%
  Fanyu Zhu$^{1,2,3}$\quad
  Jiahui An$^{1,2,3,4}$\quad
  Ni Ji$^{1,2,3,}$\thanks{Corresponding author.} \\
  \textsuperscript{1}Beijing Institute for Brain Research, \\ Chinese Academy of Medical Sciences \& 
  Peking Union Medical College\\
  \textsuperscript{2}Chinese Institute for Brain Research, Beijing\\
  \textsuperscript{3}Beijing Key Laboratory of Brain Science and Brain-Machine Interface\\
  \textsuperscript{4}State Key Laboratory of Cognitive Neuroscience and Learning,
  Beijing Normal University\\
  \texttt{\{zhufanyu,anjiahui,niji\}@cibr.ac.cn}
}

\begin{document}

\maketitle

\begin{abstract}
Efficient exploration in high-dimensional decision spaces remains a central challenge for decision-making systems. Humans, in contrast, can navigate large decision spaces with remarkable efficiency. Recent behavioral studies suggest that humans reduce dimensionality in large decision spaces by probing candidate feature dimensions, identifying reward-relevant ones, and restricting the effective decision space. Inspired by this mechanism, we propose \textbf{TDGE} (\emph{Task-Dimension-Guided Exploration}), a human-inspired, model-agnostic algorithm with an automatically constructed task-dimension--feature--item hierarchy. TDGE follows a top-down exploration strategy: it first selects task-relevant feature dimensions, then identifies informative features within those dimensions, and finally recommends concrete items based on the selected features. Experiments on MovieLens-20M, Last.fm, and Amazon recommendation datasets show that TDGE substantially improves exploration efficiency and cold-start adaptation over baseline algorithms. Comparisons with other structured algorithms and ablation studies attribute these gains to TDGE's hierarchical structure and semantic feature-space exploration, with robust results across clustering methods and hierarchy depths. Recommendation-trajectory visualizations also show exploration patterns similar to human dimension-guided behavior. 
Code is available at \url{https://github.com/LordSlime/Human-inspired-Task-Dimension-Guided-Exploration-for-Efficient-Learning}.
\end{abstract}

\section{Introduction}

Exploration in large decision spaces remains a central challenge in reinforcement learning and other sequential decision-making settings. Classical strategies such as $\epsilon$-greedy \citep{watkins1989learning, sutton1998reinforcement}, upper confidence bound (UCB) \citep{auer2002finite}, and Thompson sampling \citep{thompson1933likelihood} provide established approaches to balancing exploration and exploitation. However, efficient exploration becomes challenging when the action space is large and reward-relevant structure is unknown. Agents may spend many interactions evaluating low-reward actions while learning the context--reward relationship under partial feedback. Work on high-dimensional Bayesian optimization suggests that exploiting structured low-dimensional decompositions can improve search efficiency \citep{kandasamy2015high, rolland2018high}. These findings motivate exploration strategies that identify and exploit task-relevant structure in large decision spaces.

Humans often navigate large decision spaces efficiently. For example, when choosing a movie, a person may first narrow the search along abstract feature dimensions, such as genre, mood, or period, and then select specific titles within the reduced space. Recent behavioral work \citep{CCNPaper} suggests that humans actively reduce the effective dimensionality of a task by probing candidate feature dimensions, identifying those predictive of reward, and restricting subsequent exploration accordingly. Applying this principle to real-world decision problems requires constructing meaningful feature dimensions and learning their relevance from feedback.

This problem is especially important in online recommendation, where an agent selects items from a large catalog for each arriving user and receives click or rating feedback as reward. This setting is naturally formulated as a contextual bandit \citep{li2010contextual}, with neural variants extending reward estimation to nonlinear context--reward relationships \citep{zhou2020neural, zhang2021neural}. A further challenge is the user cold-start problem, where the system must adapt to new users with limited interaction feedback \citep{schein2002methods, panda2022approaches}. Together, online learning and user cold-start adaptation provide a practical setting for evaluating exploration methods in large decision spaces.

Prior work has explored various approaches to exploration in large decision spaces. Representative examples in reinforcement learning include continuous action embeddings with nearest-neighbor search \citep{dulac2015deep}, action elimination using auxiliary environmental signals \citep{zahavy2018learn}, and policy learning in a learned low-dimensional action space to enable
generalization across similar actions \citep{chandak2019learning}. In contextual bandits, CoFineUCB \citep{yue2012hierarchical} uses a coarse parameter subspace to guide learning in the full linear reward-parameter space, while H$_2$N-Bandit \citep{Bi-level_Hierarchical} filters candidate arms through a predefined item-category hierarchy. These methods illustrate different ways of incorporating structure into exploration. TDGE constructs semantic dimensions from item-descriptive features and treats dimensions and features as explicit intermediate actions for exploration and candidate generation.

Motivated by human dimension-level exploration, we propose \textbf{TDGE} (\emph{Task-Dimension-Guided Exploration}), a model-agnostic exploration framework for contextual-bandit recommendation. TDGE automatically groups item-descriptive features into semantic task dimensions using clustering and a KGS-based selection criterion \citep{kelley1996automated}. It then organizes exploration in a top-down manner: selecting task-relevant dimensions, selecting features within those dimensions, and recommending an item from the induced candidate pool. Selected feature scores and item--feature relevance jointly determine candidate ranking. The observed item reward updates the item-level agent and is propagated to matched selected features and their source dimensions through relevance-weighted updates. This coupling of hierarchical selection, candidate generation, and credit assignment enables feedback-driven exploration in the semantic feature space.

Our contributions are as follows.
\begin{itemize}
    \item Inspired by \textbf{human dimension-level exploration}, we propose \textbf{TDGE}, a \textbf{model-agnostic} framework that makes explicit online dimension--feature--item decisions over a semantic hierarchy \textbf{automatically constructed from item metadata}, without substantially increasing inference time. Feature-level scores guide candidate generation, while item feedback updates matched features and their source dimensions.

    \item We evaluate TDGE with \textbf{three contextual-bandit backbones}
    on MovieLens-20M, Last.fm, and Amazon. TDGE significantly reduces both online and cold-start adaptation regret on all three datasets, demonstrating \textbf{high sample efficiency} and \textbf{rapid adaptation to unseen users}.

    \item Comparisons with structured contextual-bandit baselines and
    ablation studies support the effectiveness of \textbf{semantic feature-space
    exploration and dimension-level routing}. Additional experiments demonstrate that TDGE is \textbf{robust to the choice of clustering method and hierarchy depth} within the evaluated settings.
\end{itemize}

\section{Background and Related Work}
\label{sec:background}

\subsection{Human exploration in high-dimensional environments}

A substantial body of work in cognitive science has studied how humans learn and make decisions in environments with many candidate features. A recurring finding is that humans selectively attend to reward-relevant features, thereby reducing the effective dimensionality of the learning problem \citep{niv2015reinforcement, radulescu2021human, CCNPaper}. Behavioral and neuroimaging studies of multidimensional learning tasks provide evidence that attention supports the identification of reward-relevant dimensions \citep{niv2015reinforcement}. Using eye tracking and fMRI, \citet{leong2017dynamic} further demonstrated a bidirectional interaction: attention influences value estimation and updating, while feedback guides subsequent allocation of attention. These findings support the role of dimension-level selection in learning under uncertainty. TDGE draws on this principle by constructing semantic feature dimensions and learning which dimensions to explore through recommendation feedback.

\subsection{Exploration in large decision spaces}

Efficient exploration in large action spaces remains a central challenge
in reinforcement learning and bandit learning. Representative reinforcement learning approaches include continuous action embeddings with nearest-neighbor search \citep{dulac2015deep}, action representations learned from state transitions to support generalization across similar actions \citep{chandak2019learning}, and action elimination using auxiliary environmental signals \citep{zahavy2018learn}. These methods improve learning or action selection by exploiting relationships among actions or restricting the available choices.

In bandit learning, zooming algorithms exploit similarity information
over arms or context--arm pairs to adaptively refine exploration
\citep{kleinberg2008multi, slivkins2011contextual}, while hierarchical
optimistic optimization guides exploration through hierarchical
partitions of the search space \citep{bubeck2011x}. Other methods introduce structure into reward estimation or candidate selection. CoFineUCB \citep{yue2012hierarchical} uses a coarse parameter subspace to guide linear reward estimation in the full feature space. Taxonomy-based bandits exploit known relationships among arms \citep{pandey2007bandits}, and H$_2$N-Bandit \citep{Bi-level_Hierarchical} uses available item categories in a bi-level neural architecture for category and item selection. \citet{zhu2022contextual} further develop an efficient contextual-bandit algorithm for continuous, linearly structured action spaces using supervised-learning and action-optimization oracles. Bio-inspired algorithm applying a special type of stochastic continuous Hopfield network \citep{jiang2025brain} managed to conduct uncertainty-directed exploration in ways similar to humans and animals.

Related ideas appear in high-dimensional Bayesian optimization.
Representative approaches include additive models
\citep{kandasamy2015high, rolland2018high}, tree-structured additive
decompositions \citep{han2021high}, random tree-based decompositions
\citep{ziomek2023random}, and group testing to identify active variables
\citep{hellsten2023high}. These methods improve optimization efficiency
through objective-function decomposition or input-variable selection.
TDGE instead explicitly models user preferences across semantic feature
space and item space, allowing item feedback to inform preference
learning over shared features and thereby supporting sample-efficient
exploration and rapid user adaptation.

\subsection{Contextual bandits and reinforcement learning for recommendation}

Online recommendation provides a natural setting for exploration algorithms, as the agent must select items from large catalogs and adapt to users with limited feedback. Contextual bandits model each recommendation as a one-step decision and capture the exploration--exploitation trade-off \citep{li2010contextual}. LinUCB uses linear reward estimation with upper-confidence-bound exploration, while NeuralUCB and NeuralTS combine nonlinear reward models with uncertainty-based action selection \citep{zhou2020neural, zhang2021neural}. Neural approaches to Thompson sampling also include approximate Bayesian methods evaluated by \citet{riquelme2018deep}. Beyond these approaches, EE-Net \citep{ban2021ee} uses separate exploitation and exploration networks, with the latter learning potential gains relative to the current reward estimate. Bandit methods have also been studied for personalized recommendation and the user cold-start problem \citep{tang2015personalized, silva2023user}.

Reinforcement learning methods extend recommendation to sequential decisions that optimize long-term user engagement \citep{zheng2018drn, ie2019slateq, lin2023survey}. For example, SlateQ decomposes the long-term value of a recommendation slate into item-level quantities under assumptions on user choice, making learning tractable in a combinatorial action space \citep{ie2019slateq}. TDGE focuses on a complementary aspect of recommendation: organizing candidate generation through semantic dimension--feature decisions before item-level selection. We evaluate this mechanism with linear and neural contextual-bandit backbones in online learning and user cold-start adaptation.

\section{Task-Dimension-Guided Exploration}
\label{sec:method}
We begin with the behavioral observation that motivates TDGE, and then present its three main components: feature-dimension construction from item metadata, multi-level value and uncertainty estimation with standard bandit backbones, and top-down recommendation through dimension-, feature-, and item-level selection. Full mathematical details are deferred to Appendix~\ref{app:algorithm}.

\subsection{Inspiration from human exploration}
\label{sec:inspiration}

\begin{figure}[htbp]
  \centering
  \includegraphics[width=.9\linewidth]{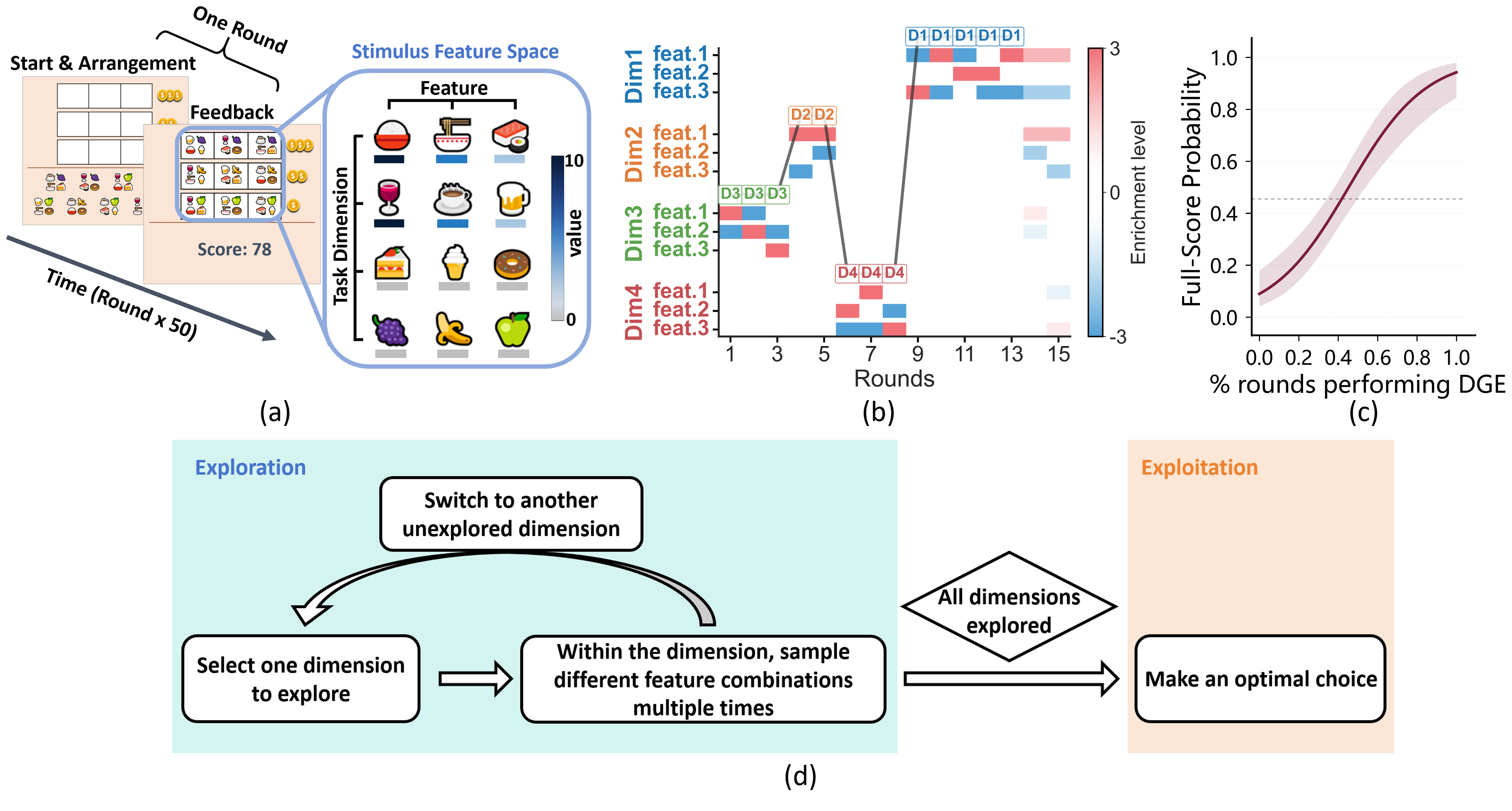}
  \caption{Human exploration through dimension-level search. (a) Behavioral task setup. (b) Reward-relevant dimensions reduce the effective action space. (c) Association between the proportion of DGE rounds and the probability of achieving a full score. (d) Schematic illustration of the observed dimension-wise exploration strategy. Adapted from An et al.~\citep{CCNPaper} with permission.}
  \label{fig:biological experiment}
\end{figure}

A recent behavioral study \citep{CCNPaper} examined how humans explore high-dimensional decision spaces using a food recommendation task. As shown in Figure~\ref{fig:biological experiment}(a), participants were asked to infer a customer's preferences by arranging food combinations to maximize reward. Each food item was described by multiple feature dimensions, but only two dimensions were reward-relevant in each session. Participants were not told which dimensions were relevant or how many dimensions mattered.

Participants adopted a structured exploration strategy consistent with dimension-guided exploration (DGE). Specifically, they first \textbf{selected one feature dimension for exploratory processing}. Within the chosen dimension, they \textbf{repeatedly sampled different feature combinations} by rearranging items across rows, allowing them to assess the relationship between that dimension and reward outcomes. After sufficiently probing one dimension, they \textbf{switched to another untested dimension} and repeated the same process. This sequential, dimension-wise exploration continued until the candidate dimensions had been evaluated, after which participants shifted from exploration to exploitation to maximize reward. Figure~\ref{fig:biological experiment}(b,d) illustrates this dimension-wise exploration process.

Figure~\ref{fig:biological experiment}(c) shows that DGE is strongly associated with task performance: as the proportion of DGE rounds increases, so does the probability of achieving a full score. This dimension-wise exploration pattern motivates TDGE, which extends the same principle into a scalable exploration framework for contextual bandit recommendation.

\subsection{Problem formulation}
\label{sec:preprocessing}

We formulate recommendation as a contextual bandit problem with partial
feedback. Let $\mathcal{I}$ denote the full item catalog. At round $t$,
the agent observes a user representation $u_t$, recommends one item
$a_t \in \mathcal{I}$, and observes reward only for that item.
For item $i$, the item-level context is
\[
x^{\mathrm{item}}_{i,t}
=
\mathrm{norm}(u_t \oplus v_i \oplus b),
\]
where $v_i$ is the item representation, $b$ is a constant bias feature,
$\oplus$ denotes concatenation, and $\mathrm{norm}(\cdot)$ denotes
$\ell_2$ normalization. The observed reward is
\[
r_t = h(x^{\mathrm{item}}_{a_t,t}) + \xi_t,
\]
where $h:\mathbb{R}^d \to [0,1]$ is an unknown expected reward function
and $\xi_t$ is zero-mean conditionally $\nu$-sub-Gaussian noise.
The objective is to minimize cumulative regret,
\[
R_T
=
\sum_{t=1}^{T}
\Bigl[
h(x^{\mathrm{item}}_{a_t^*,t})
-
h(x^{\mathrm{item}}_{a_t,t})
\Bigr],
\qquad
a_t^* \in \arg\max_{i \in \mathcal{I}}
h(x^{\mathrm{item}}_{i,t}),
\]
where $a_t^*$ is an optimal item for the current user over the full
catalog, independent of candidate sampling or filtering.
Higher-level contexts for feature dimensions and features are defined
analogously; full details and the offline candidate-sampling and reward
protocol are given in Appendix~\ref{app:problem}.

\paragraph{Dimension-guided exploration objective}
The key distinction from a standard contextual bandit lies in how
exploration is structured. A standard item-level policy explores
directly over candidate items, without explicitly restricting search
through intermediate semantic dimensions. TDGE instead first selects
a small set of feature dimensions, then selects candidate features
within those dimensions, and finally performs item-level selection
within the candidate pool $\mathcal{P}_t \subseteq \mathcal{I}$:
\[
a_t = \arg\max_{i \in \mathcal{P}_t} s^{\mathrm{item}}_{i,t},
\]
where $s^{\mathrm{item}}_{i,t}$ is the item-level selection score.
TDGE does not change the final reward objective; instead, it improves
exploration by using feature-dimension selection to construct a
task-adaptive candidate pool before item-level scoring. In this sense,
its benefit comes from better candidate-space restriction as well as
more efficient item-level search within that restricted space.

\subsection{Task-dimension extraction}
\label{sec: dimension_extraction}

TDGE requires a dimension-level exploration space that groups
item-descriptive features into semantically coherent dimensions.
Let $\mathcal{F}=\{f_1,\dots,f_M\}$ denote the feature set, such as
item tags in recommendation datasets. We seek a partition
$\mathcal{C}=\{C_1,\dots,C_{k^*}\}$ of $\mathcal{F}$, where each
cluster corresponds to one task dimension used by the dimension-level
agent.

We embed each feature $f_j$ as a normalized vector
$e_j\in\mathbb{R}^{768}$ using a pretrained Sentence-BERT model
\citep{reimers2019sentence, song2020mpnet}, and cluster the embeddings
to obtain $\mathcal{C}$. We use agglomerative hierarchical clustering
with Ward's linkage as the default method. Its dendrogram allows
feature dimensions to be extracted at different resolutions without
rerunning clustering and supports extensions with nested dimension
levels. The standard three-level TDGE framework requires only one
feature partition and can also use other clustering methods.

We determine the number of task dimensions using a KGS-based
criterion that balances cluster compactness against over-partitioning.
Each cut of the dendrogram defines a candidate partition. Among
partitions satisfying a minimum cluster-size constraint, we select
the lowest-scoring partition and denote its number of clusters by
$k^*$ (Figure~\ref{fig:kgs_dendro}(a)). Full details of the criterion
and constrained selection rule are given in
Appendix~\ref{app:clustering}. The resulting clusters capture related
semantic aspects of items, such as genre, mood, topic, or style.
Choosing a task dimension therefore directs exploration toward a
coherent group of descriptive features.
Figure~\ref{fig:kgs_dendro}(b) shows representative tag clusters
extracted from MovieLens-20M. Clustering is performed once during
preprocessing and reused throughout training and evaluation.

\begin{figure}[htbp]
    \centering
    \begin{minipage}{.38\linewidth}
        \centering
        \includegraphics[width=\linewidth]
        {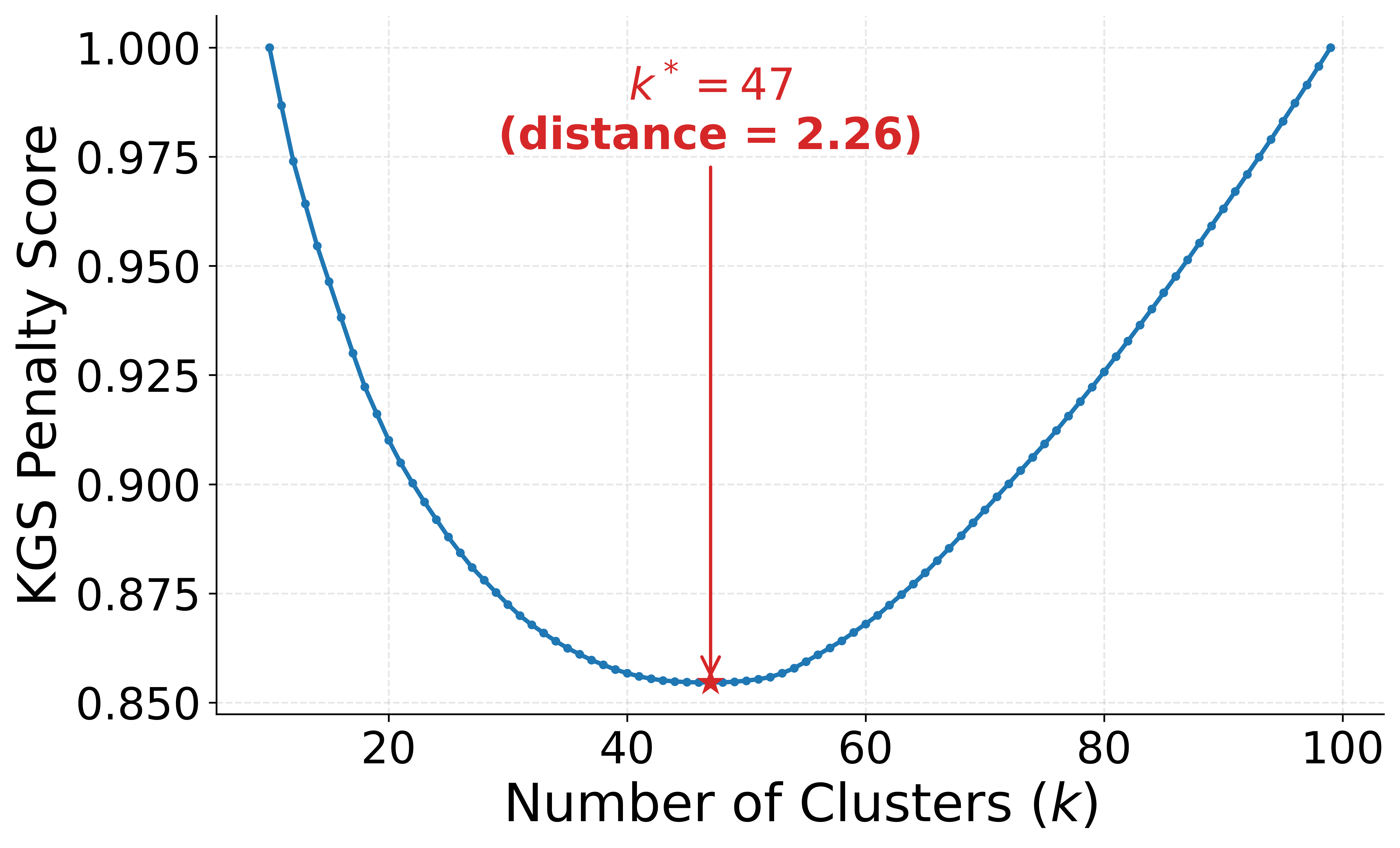}\\
        (a)
    \end{minipage}
    \hfill
    \begin{minipage}{.58\linewidth}
        \centering
        \includegraphics[width=\linewidth]
        {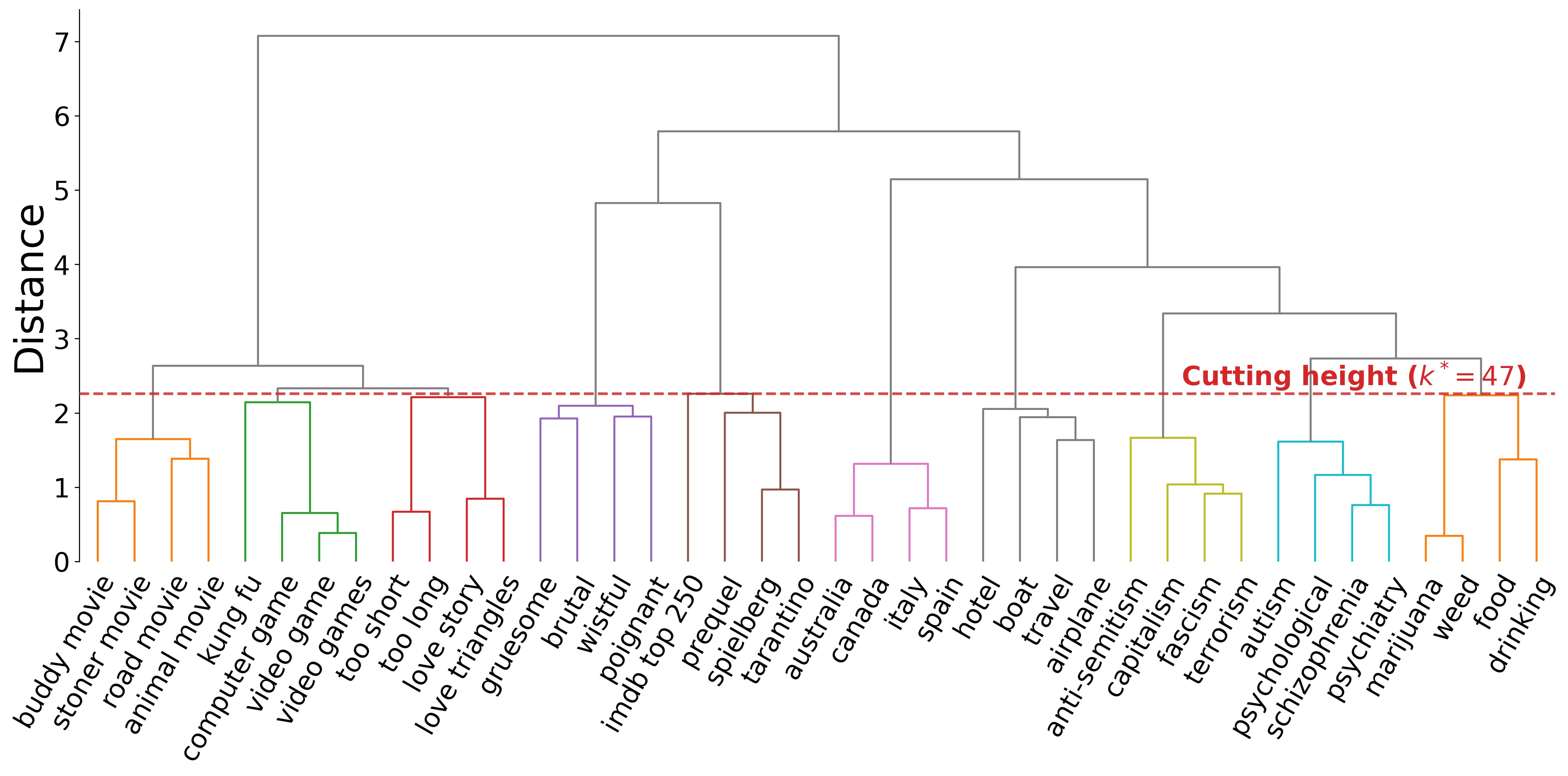}\\
        (b)
    \end{minipage}
    \caption{
    (a) Normalized KGS penalty versus cluster count on MovieLens-20M;
    $k^*$ minimizes the penalty among candidate partitions satisfying
    the cluster-size constraint.
    (b) Representative portion of the feature dendrogram:
    semantically related tags of the same color form a task dimension.
    }
    \label{fig:kgs_dendro}
\end{figure}

\subsection{Multi-Level Exploration Schedule}
\label{sec:schedule}

TDGE performs top-down exploration over three levels: task dimension,
feature, and item. The dimension-level agent identifies promising
semantic aspects of the current user's preferences and restricts
feature exploration to the selected dimensions. The feature-level
agent identifies specific attributes within those dimensions and
uses their scores to guide item candidate generation. The item-level
agent evaluates concrete items in the resulting pool and makes the
final recommendation. Each level maintains a separate contextual-bandit
agent instantiated with the same backbone.

All three levels score arms using the current user representation
and the corresponding arm representation. Let
$x_{z,t}^{(\ell)}$
be the context of arm $z$ at level
$\ell\in\{\mathrm{dim},\mathrm{feat},\mathrm{item}\}$.
The scoring and selection procedures are
\begin{equation}
\left\{
\begin{aligned}
s_{c,t}^{(\mathrm{dim})}
&= \mathcal{A}_t^{(\mathrm{dim})}
   \bigl(x_{c,t}^{(\mathrm{dim})}\bigr),
& \mathcal{D}_t
&= \operatorname{TopK}_{C_c\in\mathcal{C}}
   \bigl(s_{c,t}^{(\mathrm{dim})},K_1\bigr),\\
s_{j,t}^{(\mathrm{feat})}
&= \mathcal{A}_t^{(\mathrm{feat})}
   \bigl(x_{j,t}^{(\mathrm{feat})}\bigr),
& \mathcal{F}_t
&= \operatorname{TopK}_{f_j\in
   \bigcup_{C_c\in\mathcal{D}_t}C_c}
   \bigl(s_{j,t}^{(\mathrm{feat})},K_2\bigr),\\
s_{i,t}^{(\mathrm{item})}
&= \mathcal{A}_t^{(\mathrm{item})}
   \bigl(x_{i,t}^{(\mathrm{item})}\bigr),
& a_t
&\in \arg\max_{i\in\mathcal{P}_t}
   s_{i,t}^{(\mathrm{item})}.
\end{aligned}
\right.
\label{eq:multilevel_selection}
\end{equation}
Here, $\mathcal{A}_t^{(\ell)}$ denotes the arm-scoring rule of the
corresponding agent, incorporating reward estimation and the
backbone's exploration mechanism. $\operatorname{TopK}$ retains
up to the specified number of highest-scoring arms.
Context construction and backbone-specific scoring rules are given
in Appendices~\ref{app:problem} and~\ref{app:algorithm}.

The feature level connects semantic exploration to item selection.
Among items supplied by the candidate-sampling procedure, TDGE
ranks those sharing at least one selected feature using
\begin{equation}
S_t(i)
=
\sum_{j:\,f_j\in\mathcal{F}_t\cap\operatorname{feat}(i)}
s_{j,t}^{(\mathrm{feat})}\rho_{i,j},
\label{eq:item_filtering_score}
\end{equation}
where $\operatorname{feat}(i)$ denotes the descriptive features of
item $i$ and $\rho_{i,j}$ is its relevance to feature $f_j$.
The top $K$ matching items form $\mathcal{P}_t$, from which the
item-level agent selects $a_t$ according to
Eq.~\ref{eq:multilevel_selection}.

\paragraph{Hierarchical update}
After observing reward $r_t$, TDGE updates the item agent with
unit sample weight and propagates feedback through the matched
features
\begin{equation}
\mathcal{H}_t
=
\mathcal{F}_t\cap\operatorname{feat}(a_t).
\end{equation}
Each $f_j\in\mathcal{H}_t$ updates the feature agent using $r_t$
with sample weight $\rho_{a_t,j}$. Each source dimension $C_c$
containing matched features is updated using the same reward with
weight
\begin{equation}
\omega_{a_t,c}
=
\max_{j:\,f_j\in C_c\cap\mathcal{H}_t}
\rho_{a_t,j}.
\end{equation}
The intersection determines which selected features and dimensions
receive feedback, while relevance weights determine their
contribution to learning. This connects item-level outcomes with
preference learning over shared semantic features, allowing each
interaction to inform subsequent recommendations.

The full procedure is illustrated in Figure~\ref{fig:algorithm}.
Detailed update equations and pseudocode are provided in
Appendix~\ref{app:algorithm}.

\begin{figure}[htbp]
    \centering
    \includegraphics[width=\linewidth]
    {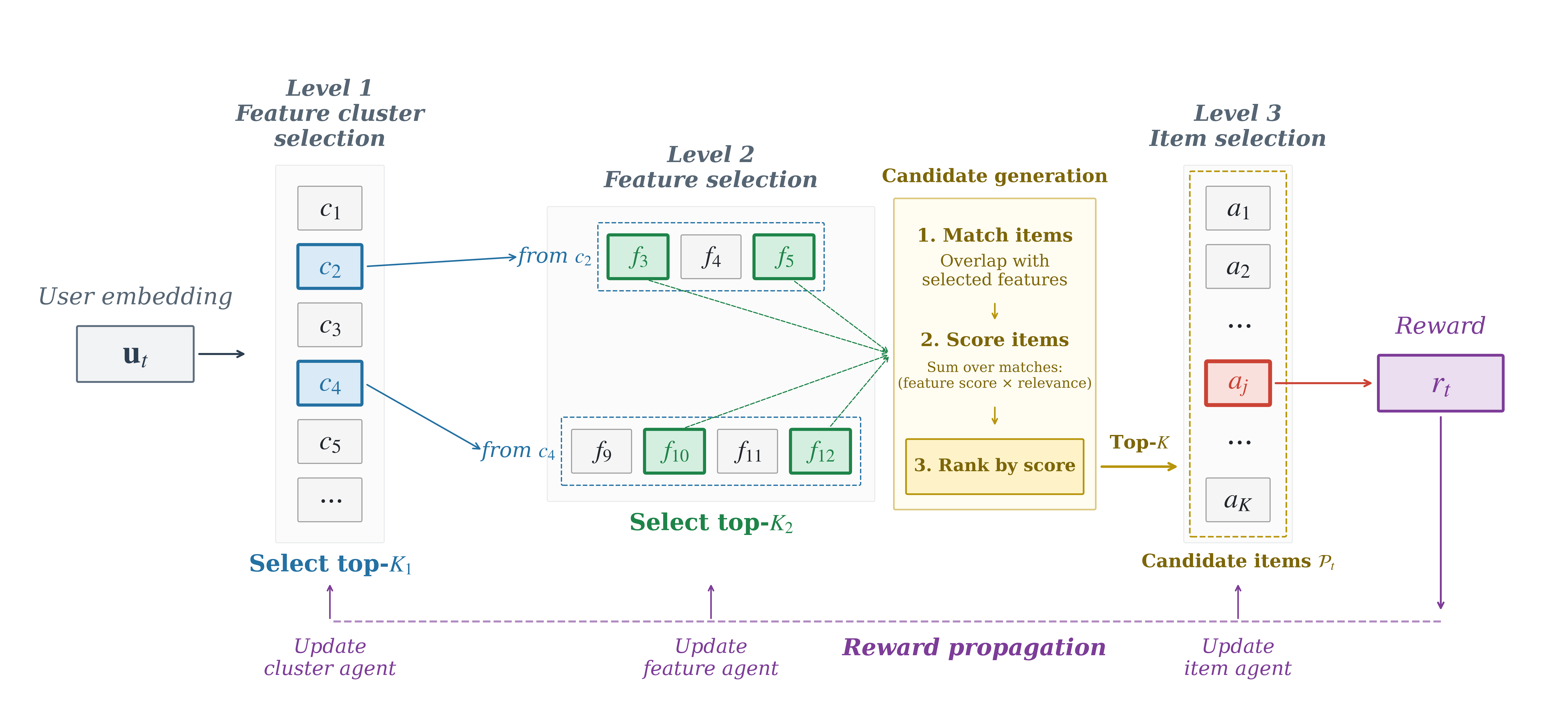}
    \caption{
    The TDGE exploration schedule. The dimension agent selects
    promising semantic aspects, the feature agent identifies
    specific attributes and guides candidate generation, and the
    item agent makes the final recommendation. The observed reward
    updates the item agent and the matched selected features and
    their source dimensions.
    }
    \label{fig:algorithm}
\end{figure}

\section{Experiments}
\label{sec:experiments}

We evaluate TDGE by addressing four questions:
(1) whether it improves sample efficiency in online learning and
cold-start adaptation across different contextual-bandit backbones;
(2) how it compares with other structured exploration methods;
(3) which components account for its gains; and
(4) how robust it is to the choice of clustering method and hierarchy depth.

\subsection{Experiment Setup}

\paragraph{Datasets}
We evaluate TDGE on MovieLens-20M, Last.fm, and the Beauty and Personal
Care subset of Amazon. For each dataset, item tags are extracted from
metadata as descriptive features. Dataset statistics, tag-selection
details, and preprocessing procedures are provided in
Appendix~\ref{app:datasets}. Users are split into an online training
pool and a user cold-start pool with a 9:1 ratio.

\paragraph{Baselines}
We compare TDGE-LinUCB, TDGE-NeuralUCB, and TDGE-NeuralTS with their
corresponding item-level backbones: LinUCB \citep{li2010contextual},
NeuralUCB \citep{zhou2020neural}, and NeuralTS \citep{zhang2021neural}.
We also compare with two structured contextual-bandit methods: CoFineUCB \citep{yue2012hierarchical}, which uses a prior
coarse parameter subspace to guide linear reward estimation, and
H$_2$N-Bandit \citep{Bi-level_Hierarchical}, which filters candidate arms
through a predefined item-category hierarchy.

\paragraph{Evaluation Protocol}
All methods follow the same offline evaluation protocol, detailed in
Appendix~\ref{app:datasets}. Within each matched TDGE--baseline pair, we
hold fixed the dataset preprocessing, user splits, item-level contexts,
interaction horizons, item-level agent configuration, and final item
budget ($K=10$); only the exploration and candidate-generation mechanism
differs. We report online cumulative regret (CReg) and cold-start
final-round regret as mean $\pm$ standard deviation over five seeds,
with significance assessed using two-sided Welch $t$-tests:$^{\ast}p<0.05$, $^{\ast\ast}p<0.01$, and $^{\ast\ast\ast}p<0.001$; n.s.\ denotes $p\geq0.05$.

\subsection{Online and Cold-Start Recommendation Results}
\label{sec:recommendation_results}

For online evaluation, each method interacts with training users for
$T=10{,}000$ rounds. For cold-start evaluation, we randomly sample 100
held-out users, initialize them with the same mean training-user
representation, and run an independent 10-step interaction trajectory for
each user.


Table~\ref{tab:main_results} summarizes both settings. With neural
backbones, TDGE significantly reduces online cumulative regret by
4.5\%--31.0\% across all six dataset--backbone comparisons,
demonstrating improved online sample efficiency. TDGE also lowers
final-round cold-start regret by 10.5\%--74.9\% in all six comparisons,
with statistically significant gains in five. With LinUCB backbone, TDGE significantly reduces online cumulative regret and final-round cold-start regret on MovieLens-20M by 31.6\% and 65.8\%, respectively, while differences on Last.fm and Amazon are not statistically significant. The limited gains on these datasets may reflect the linear reward model's restricted ability to capture complex user--item interactions, limiting the benefit of TDGE's exploration framework. These results show that TDGE's semantic hierarchy and learned routing policy support more sample-efficient exploration and adaptation to unseen users without user-specific historical contexts, with the most consistent gains obtained using neural reward models.

\begin{table}[h]
\centering
\small
\setlength{\tabcolsep}{4pt}
\caption{Online cumulative regret at $T=10{,}000$ and final-round
cold-start regret at $t=10$ for TDGE and its corresponding item-level
baselines. Lower is better. Results are mean $\pm$ sample standard
deviation over five seeds. Superscripts compare TDGE with its
corresponding backbone using two-sided Welch $t$-tests.}
\label{tab:main_results}
\resizebox{\textwidth}{!}{%
\begin{tabular}{lllcc}
\toprule
\textbf{Metric} & \textbf{Dataset} & \textbf{Backbone} &
\textbf{Baseline} & \textbf{TDGE} \\
\midrule
\multirow{9}{*}{Online CReg $\downarrow$}
& \multirow{3}{*}{MovieLens-20M}
& LinUCB
& $2299.697{\pm}20.439$
& $\mathbf{1573.142{\pm}43.499}^{\ast\ast\ast}$ \\
& & NeuralUCB
& $2417.798{\pm}15.161$
& $\mathbf{1669.303{\pm}91.162}^{\ast\ast\ast}$ \\
& & NeuralTS
& $2418.077{\pm}15.964$
& $\mathbf{1705.542{\pm}52.769}^{\ast\ast\ast}$ \\
\cmidrule(lr){2-5}
& \multirow{3}{*}{Last.fm}
& LinUCB
& $\mathbf{2438.154{\pm}43.877}$
& $2453.269{\pm}27.544\ \mathrm{n.s.}$ \\
& & NeuralUCB
& $2233.387{\pm}13.230$
& $\mathbf{1960.563{\pm}57.620}^{\ast\ast\ast}$ \\
& & NeuralTS
& $2234.352{\pm}16.528$
& $\mathbf{1975.365{\pm}81.538}^{\ast\ast}$ \\
\cmidrule(lr){2-5}
& \multirow{3}{*}{Amazon}
& LinUCB
& $1314.280{\pm}30.398$
& $\mathbf{1303.841{\pm}29.050}\ \mathrm{n.s.}$ \\
& & NeuralUCB
& $1245.925{\pm}22.689$
& $\mathbf{1169.567{\pm}32.945}^{\ast\ast}$ \\
& & NeuralTS
& $1233.443{\pm}30.148$
& $\mathbf{1177.886{\pm}26.246}^{\ast}$ \\
\midrule
\multirow{9}{*}{Cold-start Regret ($t=10$) $\downarrow$}
& \multirow{3}{*}{MovieLens-20M}
& LinUCB
& $0.234{\pm}0.004$
& $\mathbf{0.080{\pm}0.013}^{\ast\ast\ast}$ \\
& & NeuralUCB
& $0.239{\pm}0.013$
& $\mathbf{0.060{\pm}0.015}^{\ast\ast\ast}$ \\
& & NeuralTS
& $0.236{\pm}0.014$
& $\mathbf{0.078{\pm}0.026}^{\ast\ast\ast}$ \\
\cmidrule(lr){2-5}
& \multirow{3}{*}{Last.fm}
& LinUCB
& $0.260{\pm}0.006$
& $\mathbf{0.260{\pm}0.005}\ \mathrm{n.s.}$ \\
& & NeuralUCB
& $0.239{\pm}0.007$
& $\mathbf{0.214{\pm}0.007}^{\ast\ast\ast}$ \\
& & NeuralTS
& $0.240{\pm}0.010$
& $\mathbf{0.212{\pm}0.024}\ \mathrm{n.s.}$ \\
\cmidrule(lr){2-5}
& \multirow{3}{*}{Amazon}
& LinUCB
& $\mathbf{0.137{\pm}0.023}$
& $0.140{\pm}0.028\ \mathrm{n.s.}$ \\
& & NeuralUCB
& $0.099{\pm}0.012$
& $\mathbf{0.065{\pm}0.022}^{\ast}$ \\
& & NeuralTS
& $0.104{\pm}0.010$
& $\mathbf{0.061{\pm}0.012}^{\ast\ast\ast}$ \\
\bottomrule
\end{tabular}%
}
\end{table}

\subsection{Comparison with Structured Exploration Methods}
\label{sec:structured_baselines}

CoFineUCB structures exploration in the reward-parameter space, while
H$_2$N-Bandit uses a predefined item-category hierarchy. TDGE instead
explores automatically constructed semantic feature dimensions.
We compare CoFineUCB with LinUCB and TDGE-LinUCB as linear methods,
and H$_2$N-Bandit with NeuralUCB and TDGE-NeuralUCB as neural methods.
We evaluate these methods on MovieLens-20M using the same protocol
as in Section~\ref{sec:recommendation_results}.

As shown in Table~\ref{tab:structured_baselines}, TDGE-LinUCB reduces
online cumulative regret and final-round cold-start regret relative to
CoFineUCB by 30.7\% and 66.4\%, respectively. TDGE-NeuralUCB reduces the
corresponding regret relative to H$_2$N-Bandit by 29.9\% and 74.6\%.
All four comparisons are significant at $p<10^{-4}$, showing that TDGE
also improves over methods pursuing related structured-exploration
objectives.

\begin{table}[h]
\centering
\small
\setlength{\tabcolsep}{3pt}
\caption{Online cumulative regret at $T=10{,}000$ and final-round
cold-start regret at $t=10$ on MovieLens-20M. Lower is better. Results
are mean $\pm$ standard deviation over five seeds. Superscripts compare
TDGE-LinUCB with CoFineUCB and TDGE-NeuralUCB with H$_2$N-Bandit using
two-sided Welch $t$-tests.}
\label{tab:structured_baselines}
\resizebox{\textwidth}{!}{%
\begin{tabular}{lcccccc}
\toprule
\textbf{Metric} &
\textbf{LinUCB} &
\textbf{CoFineUCB} &
\textbf{TDGE-LinUCB} &
\textbf{NeuralUCB} &
\textbf{H$_2$N-Bandit} &
\textbf{TDGE-NeuralUCB} \\
\midrule
Online CReg $\downarrow$
& $2299.697{\pm}18.281$
& $2271.580{\pm}34.295$
& $\mathbf{1573.142{\pm}38.906}^{\ast\ast\ast}$
& $2417.798{\pm}13.561$
& $2379.700{\pm}17.122$
& $\mathbf{1669.303{\pm}81.538}^{\ast\ast\ast}$ \\
Cold-start Regret ($t=10$) $\downarrow$
& $0.234{\pm}0.004$
& $0.237{\pm}0.010$
& $\mathbf{0.080{\pm}0.011}^{\ast\ast\ast}$
& $0.239{\pm}0.011$
& $0.235{\pm}0.008$
& $\mathbf{0.060{\pm}0.014}^{\ast\ast\ast}$ \\
\bottomrule
\end{tabular}%
}
\end{table}

\subsection{Ablation Studies}
\label{sec:exp_ablation}

All ablations are conducted on MovieLens-20M with the five-run protocol used
for the main experiments. 

\subsubsection{Semantic Feature-Space Structure}
\label{app:ablation_feature_space}

We compare TDGE's feature-space hierarchy with a two-level item-space
alternative that clusters semantic item embeddings and selects
item-cluster--item routes. We reconstruct the feature dimensions with
Ward hierarchical clustering, $k$-means, spectral clustering, and Gaussian
mixture models. Each feature-clustering method operates on the same tag
embeddings and uses the constrained KGS criterion to choose its number of
dimensions.

\begin{table}[h]
\centering
\caption{Feature-space and clustering-method ablations on MovieLens-20M.
Lower regret is better. The superscript indicates that hierarchical feature-space TDGE outperforms
the corresponding item-clustering alternative under a two-sided Welch
$t$-test.}
\label{tab:app_clustering_ablation}
\resizebox{\textwidth}{!}{%
\begin{tabular}{llccccc}
\toprule
Metric & Backbone & Item clustering & Hierarchical & $k$-means & Spectral & GMM \\
\midrule
\multirow{3}{*}{Online CReg}
& LinUCB
& $2101.500\pm37.535^{\ast\ast\ast}$
& $1573.142\pm38.906$
& $1608.404\pm49.511$
& $1639.962\pm135.799$
& $1558.282\pm63.691$ \\
& NeuralUCB
& $2223.900\pm45.983^{\ast\ast\ast}$
& $1669.303\pm81.538$
& $1839.836\pm124.657$
& $1854.616\pm42.444$
& $1786.975\pm35.025$ \\
& NeuralTS
& $2295.800\pm13.952^{\ast\ast\ast}$
& $1705.542\pm47.198$
& $1815.254\pm112.595$
& $1792.321\pm180.387$
& $1709.182\pm94.690$ \\
\midrule
\multirow{3}{*}{Cold-start Regret ($t=10$)}
& LinUCB
& $0.219\pm0.011^{\ast\ast\ast}$
& $0.080\pm0.011$
& $0.127\pm0.005$
& $0.118\pm0.016$
& $0.133\pm0.024$ \\
& NeuralUCB
& $0.210\pm0.014^{\ast\ast\ast}$
& $0.060\pm0.014$
& $0.075\pm0.011$
& $0.072\pm0.015$
& $0.072\pm0.015$ \\
& NeuralTS
& $0.212\pm0.018^{\ast\ast\ast}$
& $0.078\pm0.023$
& $0.077\pm0.007$
& $0.065\pm0.007$
& $0.069\pm0.014$ \\
\bottomrule
\end{tabular}%
}
\end{table}

Feature-space TDGE substantially
outperforms item-space clustering for every backbone and metric, while the
four feature-clustering algorithms yield comparable results. The gains
therefore arise from exploring the semantic feature space and do not depend
on a particular clustering method.

\subsubsection{Hierarchy Depth}
\label{app:ablation_depth}

The main model uses the three-level route
\emph{fine dimension $\rightarrow$ feature $\rightarrow$ item}. We compare it
with a four-level route,
\emph{coarse dimension $\rightarrow$ fine dimension $\rightarrow$ feature
$\rightarrow$ item}. Both dimension levels are obtained from the same feature
dendrogram. The four-level model retains two coarse dimensions, five fine
dimensions, ten features, and at most ten items.

\begin{table}[htbp]
\centering
\caption{Online cumulative regret and cold-start final-round regret for the
hierarchy-depth ablation on MovieLens-20M. Lower is better.}
\label{tab:app_depth_ablation}
\begin{tabular}{llcc}
\toprule
Metric & Backbone & TDGE (3 levels) & TDGE (4 levels) \\
\midrule
\multirow{3}{*}{Online CReg}
& LinUCB
& $1573.142\pm38.906$
& $1586.160\pm69.115$ \\
& NeuralUCB
& $1669.303\pm81.538$
& $1768.740\pm142.438$ \\
& NeuralTS
& $1705.542\pm47.198$
& $1606.220\pm90.449$ \\
\midrule
\multirow{3}{*}{Cold-start Regret ($t=10$)}
& LinUCB
& $0.080\pm0.011$
& $0.090\pm0.022$ \\
& NeuralUCB
& $0.060\pm0.014$
& $0.078\pm0.017$ \\
& NeuralTS
& $0.078\pm0.023$
& $0.060\pm0.017$ \\
\bottomrule
\end{tabular}
\end{table}

None of the differences between three- and four-level TDGE is statistically
significant under two-sided Welch $t$-tests over five runs. Adding a coarse
dimension level therefore provides no consistent improvement on
MovieLens-20M. We use the three-level hierarchy because it preserves explicit
dimension--feature--item semantics with lower routing and inference
complexity. The effective hierarchy depth may depend on the semantic
structure of the dataset, and deeper hierarchies may be beneficial for data
with more complex multi-level structure.

\subsection{TDGE exhibits human-like dimension-persistent probing}
\label{sec: human-like exploration}

The behavioral study of An et al.~\citep{CCNPaper} found that participants often explored several features within one task dimension before switching to another. We examine whether TDGE produces a similar trajectory pattern in cold-start recommendation, without claiming equivalence to human cognitive mechanisms.

\begin{figure}[htbp]
  \centering
  \begin{minipage}[t]{.32\linewidth}
    \centering
    \includegraphics[width=\linewidth,height=3cm]{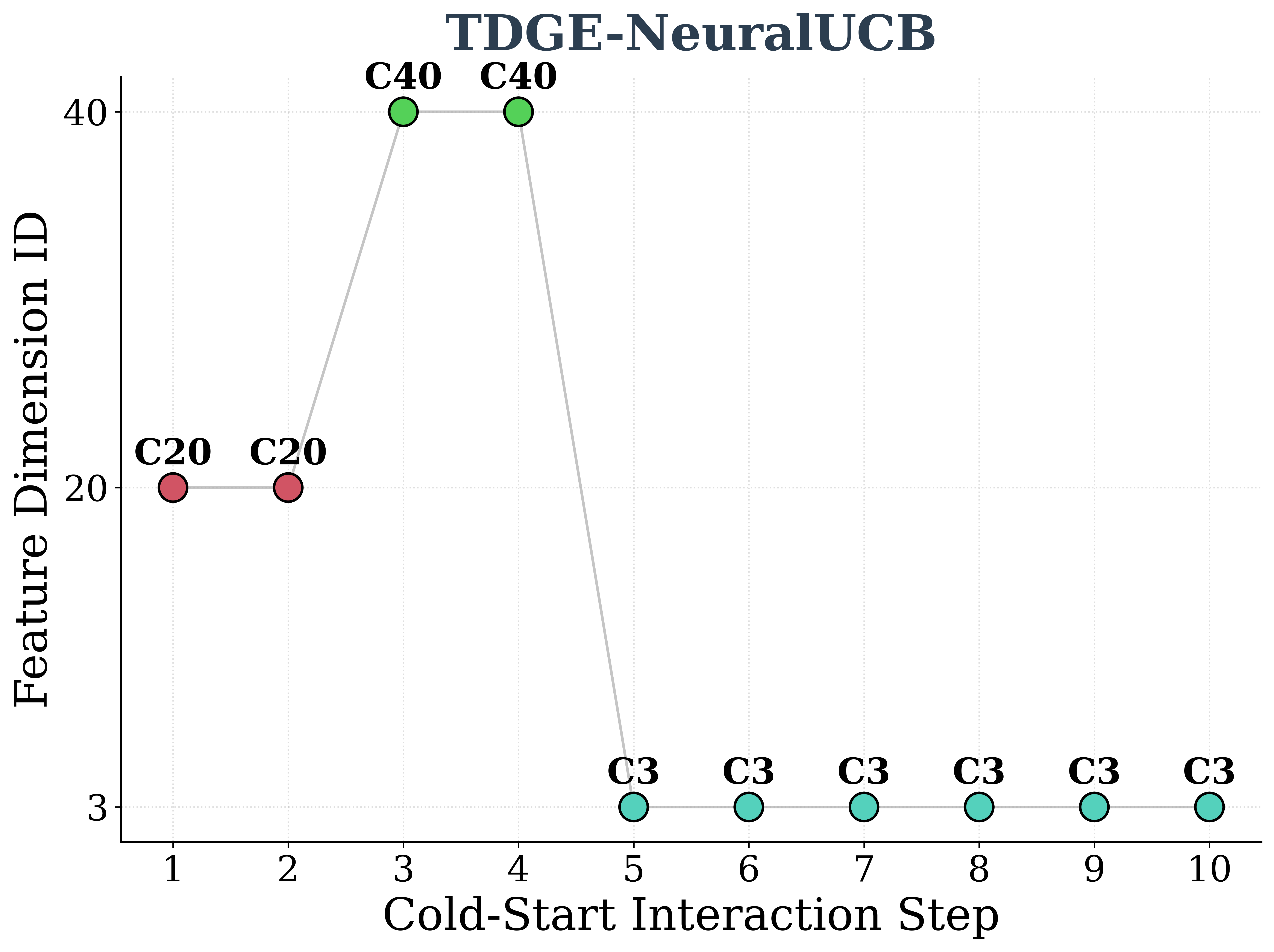}\\
    (a)
  \end{minipage}\hfill
  \begin{minipage}[t]{.32\linewidth}
    \centering
    \includegraphics[width=\linewidth,height=3cm]{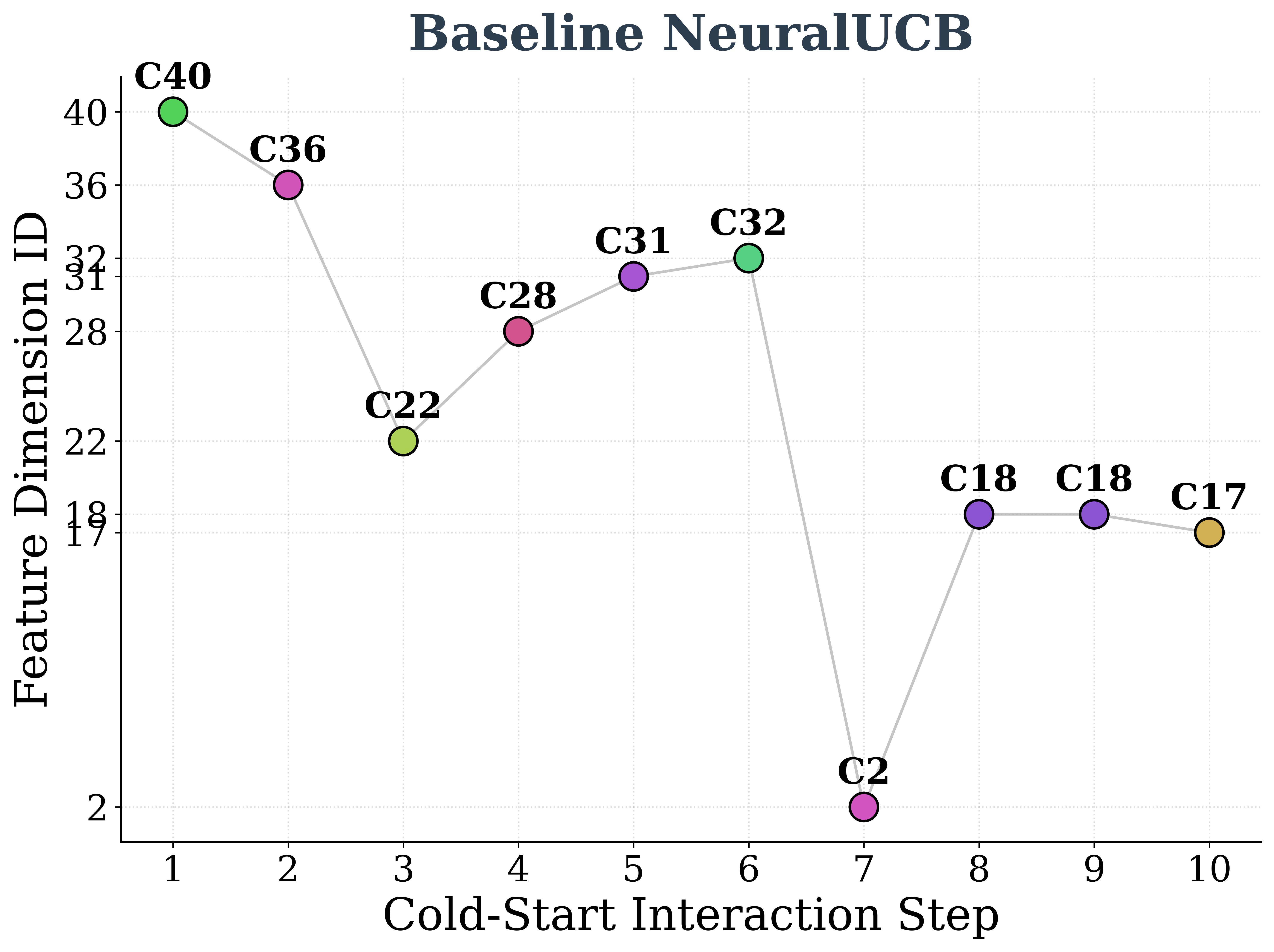}\\
    (b)
  \end{minipage}\hfill
  \begin{minipage}[t]{.32\linewidth}
    \centering
    \includegraphics[width=\linewidth]{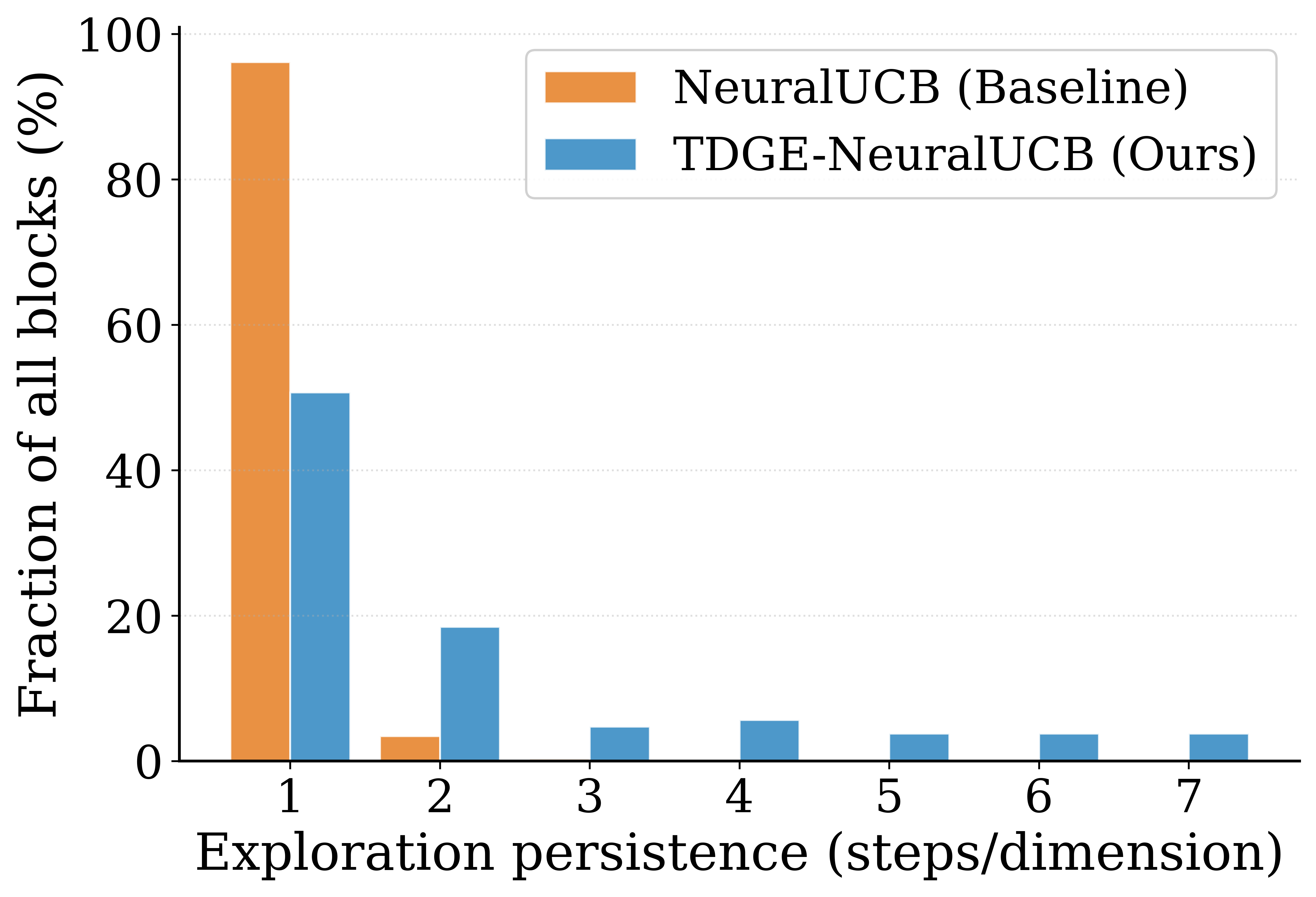}\\
    (c)
  \end{minipage}
  \caption{Feature-dimension trajectories over 10 cold-start interactions and the distribution of consecutive same-dimension run lengths on MovieLens-20M. (a) TDGE-NeuralUCB. (b) NeuralUCB. Each step in (a,b) is represented by the dominant feature dimension of the recommended item. (c) Same-dimension run-length distributions aggregated across cold-start users.}
  \label{fig:dimension_selection}
\end{figure}

Figure~\ref{fig:dimension_selection}(a,b) compares TDGE-NeuralUCB and NeuralUCB for the same cold-start user. Because the baseline does not explicitly select dimensions, both trajectories are represented by the dominant feature dimension of the final recommended item, determined by majority vote over its tags. TDGE exhibits more persistent within-dimension exploration, whereas the baseline switches more frequently.

We quantify this behavior using the run lengths of consecutive same-dimension recommendations across all MovieLens-20M cold-start users. Figure~\ref{fig:dimension_selection}(c) shows that TDGE-NeuralUCB places more probability mass on run lengths 2--5, while NeuralUCB is concentrated at length 1. Thus, TDGE produces exploration trajectories resembling the dimension-persistent pattern observed in human exploration.

\section{Conclusion}
\label{sec: conclusion and limitation}

We presented TDGE, a model-agnostic exploration framework inspired by
human dimension-level exploration. TDGE explicitly models user
preferences at the semantic dimension, feature, and item levels,
coupling exploration in semantic feature space with selection in
item space. Experiments on MovieLens-20M, Last.fm, and Amazon show significant reductions in online cumulative regret and cold-start
final-round regret, particularly with the neural backbones. Ablations support the effectiveness of semantic feature-space exploration and dimension-level routing, and demonstrate robustness to different clustering methods.

However, the current version of TDGE still has several limitations. First, the retention sizes $K_1$ and $K_2$ are fixed. Ideally, they could be
selected adaptively or learned online to better match different
datasets and user regimes. Second, the limited effect of hierarchy
depth in our ablations may be due to the structure of the evaluated
data. Experiments on datasets with more complex hierarchical
structure could help determine when deeper hierarchies are
beneficial. Third, we evaluate TDGE only with contextual bandit
backbones; extending the same dimension-guided exploration principle to
general reinforcement learning is an important direction for
future work.

In conclusion, these findings support explicit exploration over semantic dimensions as an effective way to improve sample efficiency in contextual-bandit recommendation.

\section{Acknowledgments}
This study was funded by Project (2025ZD0217400) supported by Brain Science and Brain-Like Intelligence Technology —National Science and Technology Major Project; and the CAMS Innovation Fund for Medical Sciences (CIFMS), 2024-RC180-02. We acknowledge Beijing Key Laboratory of Brain Science and Brain-Machine Interface; and Fundamental and Interdisciplinary Disciplines Breakthrough Plan of the Ministry of Education of China (JYB2025XDXM504).

\clearpage

{\small
\bibliographystyle{plainnat}
\bibliography{refs}
}

\clearpage
\appendix


\section{Full Problem Formulation, Exploration Schedule, and Evaluation Protocol}
\label{app:problem}

\subsection{Recommendation Formulation and Context Construction}

We formulate recommendation as a contextual-bandit problem with partial
feedback. Let $\mathcal{I}$ be the item catalog and $\mathcal{U}$ the user
set. At round $t$, the agent observes a user representation
$u_t\in\mathbb{R}^{50}$, recommends one item $a_t$, and observes feedback only
for that item.

For offline evaluation, let $\mathcal{I}^{\mathrm{eval}}_t=\mathcal{I}^{+}_t$
contain the items with recorded feedback for the current user. We do not add
unobserved user--item pairs to this set or impute rewards for missing
feedback. Before selecting an item, an agent observes the item contexts but
not their logged rewards; after selecting $a_t$, only its normalized logged
reward is revealed. Both the flat baselines and TDGE make the final decision
over at most $K=10$ item arms. Items remain available after selection, as in
the standard repeated-arm contextual-bandit setting.    

The representation of an action $z$---a feature dimension, feature, or
item---is denoted by $v_z\in\mathbb{R}^{50}$. Its contextual representation is
\begin{equation}
    x_{z,t}
    =
    \operatorname{norm}
    \left(u_t\oplus v_z\oplus 0.01\right)
    \in\mathbb{R}^{101},
    \label{eq:app_rec_context}
\end{equation}
where $\oplus$ denotes concatenation and $\operatorname{norm}$ denotes
$\ell_2$ normalization. Item and feature representations are obtained by
projecting their semantic embeddings to 50 dimensions. A dimension
representation is the normalized centroid of the projected features assigned
to that dimension. This common context construction allows the same
contextual-bandit backbone to operate at every TDGE level.

\subsection{Multi-Level Selection and Candidate Construction}

Let $\mathcal{C}$ be the set of feature dimensions and $C_c$ the features in
dimension $c$. At round $t$, the dimension-level agent retains
\begin{equation}
    \mathcal{D}_t
    =
    \operatorname{TopK}_{c\in\mathcal{C}}
    \left(s_{c,t}^{(\mathrm{dim})},K_1\right).
\end{equation}
The feature-level agent then forms
\begin{equation}
    \mathcal{G}_t=\bigcup_{c\in\mathcal{D}_t}C_c
\end{equation}
and retains
\begin{equation}
    \mathcal{F}_t
    =
    \operatorname{TopK}_{j\in\mathcal{G}_t}
    \left(s_{j,t}^{(\mathrm{feat})},K_2\right).
\end{equation}
Let $q_{t,j}=s_{j,t}^{(\mathrm{feat})}$ be the score of selected feature $j$,
$\rho_{i,j}$ the relevance of feature $j$ to item $i$, and
$\operatorname{feat}(i)$ the feature set of item $i$. If continuous relevance
is unavailable, $\rho_{i,j}=1$ for $j\in\operatorname{feat}(i)$. The eligible
item set is
\begin{equation}
    \mathcal{E}_t
    =
    \left\{
        i\in\mathcal{I}^{\mathrm{eval}}_t:
        \mathcal{F}_t\cap\operatorname{feat}(i)\neq\emptyset
    \right\}.
\end{equation}
TDGE scores each eligible item by
\begin{equation}
    S_t(i)
    =
    \sum_{j\in\mathcal{F}_t\cap\operatorname{feat}(i)}
    q_{t,j}\rho_{i,j},
    \label{eq:app_candidate_score}
\end{equation}
and retains
\begin{equation}
    \mathcal{P}_t
    =
    \operatorname{TopK}_{i\in\mathcal{E}_t}\left(S_t(i),K\right).
\end{equation}
If $\mathcal{E}_t$ is empty, TDGE resamples the dimension--feature route before
making a recommendation rather than introducing an item outside the selected
semantic route. The item-level bandit then selects $a_t$ from
$\mathcal{P}_t$.

\subsection{Reward, Regret, and Hierarchical Credit Assignment}

The selected item receives reward
\begin{equation}
    r_t=h(x_{a_t,t})+\xi_t,
\end{equation}
where $h:\mathbb{R}^{101}\rightarrow[0,1]$ is the unknown expected reward and
$\xi_t$ is zero-mean conditionally sub-Gaussian noise. In the offline
evaluation, the oracle is the highest-reward observed item,
\begin{equation}
    a_t^*=\arg\max_{i\in\mathcal{I}^{+}_t} h(x_{i,t}),
\end{equation}
and cumulative regret is
\begin{equation}
    R_T=\sum_{t=1}^{T}
    \left[h(x_{a_t^*,t})-h(x_{a_t,t})\right].
\end{equation}

The item agent is updated at every completed round. Credit is propagated to
the upper levels only through features shared by the selected route and the
recommended item:
\begin{equation}
    \mathcal{H}_t
    =
    \mathcal{F}_t\cap\operatorname{feat}(a_t).
    \label{eq:app_hard_intersection}
\end{equation}
Each $j\in\mathcal{H}_t$ is updated with sample weight
$\rho_{a_t,j}$, and each source dimension $c$ is updated with
\begin{equation}
    \omega_{a_t,c}
    =
    \max_{j\in\mathcal{H}_t\cap C_c}\rho_{a_t,j}.
\end{equation}
Thus, the hard intersection determines which upper-level actions receive
credit, while $\rho$ provides graded credit among matched actions. For
MovieLens, $\rho$ is the genome relevance; for Last.fm and Amazon it is one.
Because the item pool contains only items matched by at least one selected
feature, $\mathcal{H}_t$ is nonempty for every completed recommendation round.

\section{Task-Dimension Extraction with KGS-Based Clustering}
\label{app:clustering}

Let $\mathcal{F}=\{f_1,\ldots,f_M\}$ be the descriptive feature set and
$e_j\in\mathbb{R}^{768}$ the normalized Sentence-BERT embedding of feature
$f_j$. Ward hierarchical clustering produces a dendrogram over
$\mathcal{F}$. A cut yielding $k$ clusters defines
$\Pi_k=\{C_1^{(k)},\ldots,C_k^{(k)}\}$. We select $k$ with the constrained
Kelley--Gardner--Sutcliffe (KGS) criterion.

For partition $\Pi_k$, its within-cluster sum of squares is
\begin{equation}
    \mathcal{W}(k)
    =
    \sum_{c=1}^{k}\sum_{f_j\in C_c^{(k)}}
    \left\|e_j-\mu_c^{(k)}\right\|_2^2,
    \qquad
    \mu_c^{(k)}
    =
    \frac{1}{|C_c^{(k)}|}\sum_{f_j\in C_c^{(k)}}e_j.
\end{equation}
Over candidate values $\mathcal{K}$, the KGS penalty is
\begin{equation}
    \operatorname{KGS}(k)
    =
    \frac{\mathcal{W}(k)-\mathcal{W}_{\min}}
         {\mathcal{W}_{\max}-\mathcal{W}_{\min}}
    +
    \frac{k-k_{\min}}{k_{\max}-k_{\min}}.
    \label{eq:app_kgs}
\end{equation}
The first term favors compact clusters and the second penalizes excessive
fragmentation. A cut is valid only if every cluster contains at least
$\eta=5$ features. The selected number of dimensions is
\begin{equation}
    k^*=\arg\min_{k\in\Omega}\operatorname{KGS}(k),
    \qquad
    \Omega=\{k\in\mathcal{K}:|C_c^{(k)}|\geq5\ \forall c\}.
\end{equation}
We search $k\in\{10,\ldots,99\}$ for all three datasets. This procedure yields 47, 37, and 35 fine-grained task dimensions for MovieLens-20M, Last.fm, and Amazon, respectively.

\section{Value Estimators and Full TDGE Procedure}
\label{app:algorithm}

TDGE uses the same top-down routing procedure with different
contextual-bandit backbones. For level
$\ell\in\{\mathrm{dim},\mathrm{feat},\mathrm{item}\}$, let $(x,r,\omega)$
denote an update observation. Item updates use $\omega=1$; matched feature
and dimension updates use the relevance weights defined above. It should be noted that our NeuralUCB and NeuralTS implementations estimate uncertainty
using penultimate-layer features rather than full-network gradients.
The same variants are used for the item-level TDGE and baselines.

\paragraph{LinUCB}
At level $\ell$, LinUCB maintains
$P_t^{(\ell)}=(A_t^{(\ell)})^{-1}$ and $b_t^{(\ell)}$, with
$\widehat\theta_t^{(\ell)}=P_t^{(\ell)}b_t^{(\ell)}$. The score of action $z$
is
\begin{equation}
    s_{z,t}^{(\ell)}
    =
    (x_{z,t}^{(\ell)})^\top\widehat\theta_t^{(\ell)}
    +\alpha_\ell
    \sqrt{(x_{z,t}^{(\ell)})^\top
    P_t^{(\ell)}x_{z,t}^{(\ell)}}.
\end{equation}
For $(x,r,\omega)$, the weighted update is
\begin{equation}
P^{(\ell)}
\leftarrow P^{(\ell)}-
\frac{\omega P^{(\ell)}xx^\top P^{(\ell)}}
{1+\omega x^\top P^{(\ell)}x},
\qquad
b^{(\ell)}\leftarrow b^{(\ell)}+\omega r x.
\end{equation}

Within each round, the feature agent is updated once per matched
feature, and the dimension agent once per matched source dimension,
with updates applied sequentially within each agent.

\paragraph{NeuralUCB}
NeuralUCB maintains a reward network $f_{\theta_t^{(\ell)}}$ and an inverse
uncertainty matrix $P_t^{(\ell)}$. Let
$\phi_{\theta_t^{(\ell)}}(x)$ be its penultimate-layer feature. The action
score is
\begin{equation}
    s_{z,t}^{(\ell)}
    =f_{\theta_t^{(\ell)}}(x_{z,t}^{(\ell)})
    +\alpha_\ell
    \sqrt{\phi_{\theta_t^{(\ell)}}(x_{z,t}^{(\ell)})^\top
    P_t^{(\ell)}
    \phi_{\theta_t^{(\ell)}}(x_{z,t}^{(\ell)})}.
\end{equation}
Observations are stored in a level-specific replay buffer and optimized with
weighted squared loss,
\begin{equation}
    \mathcal{L}^{(\ell)}(\theta)
    =\frac{1}{|\mathcal{B}^{(\ell)}|}
    \sum_{(x_\tau,r_\tau,\omega_\tau)\in\mathcal{B}^{(\ell)}}
    \omega_\tau\bigl(f_\theta(x_\tau)-r_\tau\bigr)^2.
\end{equation}
After updating the network, $P_t^{(\ell)}$ is updated by the same weighted
Sherman--Morrison rule using the new penultimate-layer feature.

\paragraph{NeuralTS}
NeuralTS uses the same reward network, replay training, and uncertainty
matrix, but samples
\begin{equation}
    s_{z,t}^{(\ell)}
    \sim\mathcal{N}\!\left(
    f_{\theta_t^{(\ell)}}(x_{z,t}^{(\ell)}),
    \alpha_\ell^2
    \phi_{\theta_t^{(\ell)}}(x_{z,t}^{(\ell)})^\top
    P_t^{(\ell)}
    \phi_{\theta_t^{(\ell)}}(x_{z,t}^{(\ell)})
    \right)
\end{equation}
and selects the action with the largest sampled value.

\begin{algorithm}[H]
\caption{Three-level TDGE used in the main recommendation experiments}
\label{alg:app_full_tdge}
\begin{algorithmic}[1]
\REQUIRE Dimensions $\mathcal{C}=\{C_1,\ldots,C_{k^*}\}$; agents
$\mathcal{A}^{(\mathrm{dim})}$, $\mathcal{A}^{(\mathrm{feat})}$,
$\mathcal{A}^{(\mathrm{item})}$; budgets $K_1,K_2,K$.
\FOR{$t=1,\ldots,T$}
    \STATE Observe $u_t$ and construct $\mathcal{I}^{\mathrm{eval}}_t$.
    \STATE Construct dimension contexts using~\eqref{eq:app_rec_context} and score them with the dimension-level agent; retain the top $K_1$ dimensions $\mathcal{D}_t$.
    \STATE Score the features in $\bigcup_{c\in\mathcal{D}_t}C_c$; retain the
    top $K_2$ features $\mathcal{F}_t$ and their scores $q_{t,j}$.
    \STATE Compute $S_t(i)$ by~\eqref{eq:app_candidate_score} for evaluation candidates having nonempty overlap with $\mathcal{F}_t$.
    \IF{at least one overlapping candidate exists}
        \STATE Retain the top $K$ scored items as $\mathcal{P}_t$.
    \ELSE
        \STATE Resample the dimension--feature route before making a
        recommendation; do not add an item outside the selected route.
    \ENDIF
    \STATE The item agent recommends
    $a_t=\arg\max_{i\in\mathcal{P}_t}s_{i,t}^{(\mathrm{item})}$ and observes
    $r_t$.
    \STATE Update the item agent with $(x_{a_t,t}^{(\mathrm{item})},r_t,1)$.
    \STATE Set $\mathcal{H}_t=\mathcal{F}_t\cap\operatorname{feat}(a_t)$.
    \FOR{$j\in\mathcal{H}_t$}
        \STATE Update the feature agent with
        $(x_{j,t}^{(\mathrm{feat})},r_t,\rho_{a_t,j})$.
    \ENDFOR
    \FOR{$c\in\mathcal{D}_t$ such that $\mathcal{H}_t\cap C_c\neq\varnothing$}
        \STATE Update the dimension agent with
        $(x_{c,t}^{(\mathrm{dim})},r_t,
        \max_{j\in\mathcal{H}_t\cap C_c}\rho_{a_t,j})$.
    \ENDFOR
\ENDFOR
\end{algorithmic}
\end{algorithm}

\section{Recommendation Datasets and Preprocessing}
\label{app:datasets}

\begin{table}[htbp]
  \centering
  \caption{Recommendation-dataset statistics after preprocessing.}
  \label{tab:app_dataset_stats}
  \resizebox{\textwidth}{!}{%
    \begin{tabular}{lrrrrrr}
      \toprule
      Dataset & \# Users & \# Items & \# Features &
      \# Dimensions ($k^\ast$) & \# Interactions & Sparsity \\
      \midrule
      MovieLens-20M & 5,000 & 10,000 & 1,128 & 47 & 5,068,810 & 89.86\% \\
      Last.fm       & 1,892 & 10,000 & 2,074 & 37 & 84,365    & 99.55\% \\
      Amazon        & 5,000 & 43,724 & 5,743 & 35 & 494,083   & 99.77\% \\
      \bottomrule
    \end{tabular}%
  }
\end{table}

\paragraph{MovieLens-20M}
We retain the 10,000 most frequently rated movies with tag-genome coverage
and the 5,000 most active users. Ratings are divided by five. For hierarchical
routing and credit assignment, each movie is associated with its ten
highest-relevance genome tags. Its semantic item representation is the
relevance-weighted average of its 50 highest-relevance tag embeddings before
PCA projection.

\paragraph{Last.fm}
We retain the 10,000 most frequently interacted artists with tag annotations
and up to 2,000 active users, resulting in 1,892 users after preprocessing.
Listening counts are transformed by $\log(1+x)$ and scaled by each user's
maximum to $[0.1,1.0]$. Tags used at most five times globally are removed.
Artist--tag relevance is binary, and up to ten associated tags define the
route and credit set of each artist. The semantic artist representation
averages up to 50 associated tag embeddings.

\paragraph{Amazon}
We use the Beauty and Personal Care subset, retain the 5,000 most active users,
and keep products with at least five interactions among those users. We
extract informative metadata key--value pairs and convert each pair into a
textual feature such as ``Skin Type: Oily.'' Features occurring on fewer than
five products are removed, and each product retains up to five features.
Ratings are divided by five, and a product representation is the unweighted
average of its retained feature embeddings.

\paragraph{Shared semantic preprocessing}
Feature strings are encoded with sentence-transformers(all-mpnet-base-v2 \citep{song2020mpnet}) and $\ell_2$ normalized.
Clustering is performed in the original 768-dimensional embedding space.
Item embeddings are then reduced to 50 dimensions with PCA; the same PCA
transform is applied to feature embeddings, and dimension arms are normalized
feature-centroid vectors in this shared space. Users are split before
factorization, and 50-dimensional truncated-SVD representations are fitted
using only the online-training users. Each evaluated cold-start user is
initialized with the same normalized mean training-user representation, so
no user-specific interaction history is included in its context.

\paragraph{Offline evaluation-environment construction.}
For each seed, retained users are split into online-training and held-out
cold-start pools with a 9:1 ratio before fitting the user representation
model. The truncated-SVD model is fitted only to the online-training users.
For cold-start evaluation, 100 held-out users are sampled, and each begins an
independent 10-step trajectory with the same normalized mean training-user
representation. The held-out users' logged interactions are used only by the
offline evaluator to define the recorded-feedback decision sets and rewards;
they are not used to fit the representation model or construct user contexts.

For user $u_t$, the evaluation decision set
$\mathcal{I}^{\mathrm{eval}}_t$ contains the items with recorded feedback for
that user. Missing user--item pairs are excluded rather than treated as
negative feedback. The logged reward of an item is hidden until that item is
selected. All compared methods use the same preprocessing, user splits,
recorded-feedback decision sets, reward definition, random seeds, interaction
horizons, and final item budget.

\paragraph{Fair comparison of candidate generation.}
All methods use the same evaluation decision sets, contexts, rewards,
interaction horizons, and final item-level budget of $K=10$. The item-level
agent in each TDGE variant has exactly the same architecture and
hyperparameters as its corresponding flat baseline. A flat baseline uniformly
samples up to $K$ items from $\mathcal{I}^{\mathrm{eval}}_t$ for item-level
scoring, whereas TDGE uses its learned dimension--feature route to construct
the $K$-item pool. Candidate generation is part of the TDGE algorithm and is
therefore included in end-to-end regret. The NoFD ablation provides a
feature-only semantic retrieval control with the same downstream retrieval,
item-level selection, and update rules, thereby isolating the additional
contribution of learned dimension-level routing.

This design directly evaluates TDGE's central advantage. In a large item
catalog, uniform sampling is unlikely to retrieve high-reward items within a
limited item-scoring budget. TDGE instead narrows the catalog through learned
user preferences at the semantic dimension and feature levels, enabling the
item-level agent to evaluate more relevant candidates and improving sample
efficiency.

\paragraph{Asset provenance and usage terms.}
MovieLens-20M is used under the GroupLens research-use terms, and the
HetRec 2011 Last.fm dataset is used under its non-commercial research
terms. Amazon Reviews 2023 is obtained from the official McAuley Lab
research release. We do not redistribute the raw third-party datasets.
The \texttt{sentence-transformers/all-mpnet-base-v2} model is released
under the Apache License 2.0. All baseline methods are properly cited.
CoFineUCB is independently implemented from its published algorithmic
description, whereas our H$_2$N-Bandit baseline is adapted from the
implementation publicly released by its original authors. All external assets are credited and used in accordance with their
applicable terms.

\section{Experimental Configuration}
\label{app:hyperparams}

All main recommendation results use five independent runs with seeds
$2026$--$2030$. Statistical significance is assessed on run-level final
metrics using two-sided Welch $t$-tests.

\begin{table}[htbp]
\centering
\caption{Shared recommendation hyperparameters.}
\label{tab:app_shared_hyper}
\begin{tabular}{lc}
\toprule
Parameter & Value \\
\midrule
Online rounds ($T$) & 10,000 \\
Independent runs & 5 \\
Final item budget ($K$) & 10 \\
Train/cold-start user split & 90\%/10\% \\
Cold-start users per run & 100 \\
Cold-start steps per user & 10 \\
Recommendation context dimension & 101 \\
Semantic embedding dimension after PCA & 50 \\
Base random seeds & 2026--2030 \\
\bottomrule
\end{tabular}
\end{table}

\paragraph{Hyperparameter selection}
Hyperparameters are selected separately for each dataset--backbone pair.
For the flat backbones, we tune the exploration coefficient $\alpha$ and
regularization parameter $\lambda$ by grid search over
\[
\alpha\in\{0.01,0.1,1,10\},
\qquad
\lambda\in\{0.01,0.1,1,10\}.
\]
For TDGE, the dimension- and feature-level values of $\alpha$ and $\lambda$
are selected from the same grids. The retention budgets are selected from
$K_1\in\{1,2,3,5,8\}$ and
$K_2\in\{4,6,10,16,24\}$.

To ensure a controlled comparison at the item level, we first tune each
flat backbone and then fix its selected $(\alpha,\lambda)$ pair for the
item-level agent of the corresponding TDGE variant. The item-level agent
also uses exactly the same architecture, optimizer, learning rate, batch
size, replay-buffer size, and update schedule as the corresponding flat
baseline. Only the dimension- and feature-level agents and their retention
budgets are specific to TDGE. Thus, performance differences cannot be
attributed to a stronger item-level reward estimator in TDGE.

\begin{table}[htbp]
\centering
\caption{Selected dimension and feature retention budgets $(K_1,K_2)$.}
\label{tab:app_schedule_hyper}
\begin{tabular}{lccc}
\toprule
Method & MovieLens-20M & Last.fm & Amazon \\
\midrule
TDGE-LinUCB    & $(5,10)$ & $(2,4)$  & $(1,6)$ \\
TDGE-NeuralUCB & $(5,10)$ & $(3,10)$ & $(3,24)$ \\
TDGE-NeuralTS  & $(5,10)$ & $(1,24)$ & $(5,24)$ \\
\bottomrule
\end{tabular}
\end{table}

\begin{table}[htbp]
\centering
\caption{Selected exploration and regularization parameters
$(\alpha,\lambda)$ for the flat backbones.}
\label{tab:app_flat_agent_hyper}
\begin{tabular}{lccc}
\toprule
Backbone & MovieLens-20M & Last.fm & Amazon \\
\midrule
LinUCB
& $(0.01,1)$
& $(0.01,1)$
& $(1,0.01)$ \\
NeuralUCB
& $(0.01,0.1)$
& $(0.01,0.01)$
& $(1,0.1)$ \\
NeuralTS
& $(0.1,10)$
& $(0.01,10)$
& $(0.1,10)$ \\
\bottomrule
\end{tabular}
\end{table}

\begin{table}[htbp]
\centering
\caption{Selected TDGE parameters. Each cell lists
$(\alpha,\lambda)_{\mathrm{dim}}/
(\alpha,\lambda)_{\mathrm{feat}}/
(\alpha,\lambda)_{\mathrm{item}}$.
For each dataset--backbone pair, the item-level parameters are identical
to those of the corresponding flat backbone in
Table~\ref{tab:app_flat_agent_hyper}.}
\label{tab:app_tdge_agent_hyper}
\resizebox{\textwidth}{!}{%
\begin{tabular}{lccc}
\toprule
Method & MovieLens-20M & Last.fm & Amazon \\
\midrule
TDGE-LinUCB
& $(1,1)/(1,1)/(0.01,1)$
& $(0.1,1)/(1,0.1)/(0.01,1)$
& $(1,1)/(0.1,1)/(1,0.01)$ \\
TDGE-NeuralUCB
& $(0.1,1)/(0.1,0.1)/(0.01,0.1)$
& $(1,10)/(0.1,10)/(0.01,0.01)$
& $(0.1,10)/(0.1,10)/(1,0.1)$ \\
TDGE-NeuralTS
& $(0.1,1)/(0.1,1)/(0.1,10)$
& $(0.1,10)/(0.1,10)/(0.01,10)$
& $(0.1,0.1)/(0.1,0.1)/(0.1,10)$ \\
\bottomrule
\end{tabular}%
}
\end{table}

\paragraph{Neural network configuration}
All neural agents use a two-hidden-layer feedforward network
$d\rightarrow128\rightarrow128\rightarrow1$ with ReLU activations,
Adam with learning rate $10^{-3}$, ten gradient steps per update, and
a replay buffer containing at most 2,000 observations. The dimension-
and feature-level agents use a batch size of 128, whereas the flat
baselines and the item-level agents of TDGE use a batch size of 64.
When the replay buffer contains fewer observations than the specified
batch size, all available observations are used. Therefore, within each
dataset--backbone pair, TDGE and its corresponding baseline use the same
item-level model and training configuration.

\paragraph{Structured baselines}
The matched MovieLens comparison includes CoFineUCB and H$_2$N-Bandit.
CoFineUCB uses a five-dimensional coarse parameter subspace learned only from
training-user preference profiles; its fine and coarse exploration
coefficients and regularization parameters match the LinUCB configuration.
H$_2$N-Bandit uses the raw MovieLens genre taxonomy, retains its top four
genre categories, and applies its neural item selector to at most ten items.
All compared methods use the same five seeds, user split, interaction
horizon, initial candidate-sampling protocol, item-level context
representations, and final item budget.

\paragraph{Compute resources}
All experiments were conducted on a single workstation equipped with a
13th Gen Intel Core i9-13900K CPU, 128\,GB of system memory, and one
NVIDIA GeForce RTX~4090 GPU with 24\,GB of memory. No distributed or
multi-GPU training was used. For an online horizon of $T=10{,}000$, a
single run required approximately one minute for LinUCB-based methods
and seven minutes for NeuralUCB- and NeuralTS-based methods. Reproducing
the complete set of reported experiments required approximately
three hours of wall-clock time. The preprocessed datasets occupy approximately 1.2 GB of storage.

\section{Additional ablation studies}
\label{app:ablation}

\subsection{Importance of the Feature-Dimension Level}
\label{app:ablation_dimension_level}

The NoFD ablation serves as a feature-only semantic candidate-generation
control. It removes the dimension-level agent while retaining the same
feature-level agent, semantic candidate scoring, item retrieval, item-level
agent, and update procedure as full TDGE. At each round, it
uniformly samples 50 features from the full feature vocabulary, applies the
same feature-level agent to retain the top $K_2$ features, and then uses the
same candidate-generation, item-selection, and update procedures as full
TDGE. 

\begin{table}[htbp]
\centering
\caption{Online cumulative regret and cold-start final-round regret for the
feature-dimension-level ablation on MovieLens-20M. Lower is better. Superscripts compare full TDGE with NoFD using two-sided Welch $t$-tests.}
\label{tab:app_no_fd_ablation}
\begin{tabular}{llcc}
\toprule
Metric & Backbone & NoFD & Full TDGE \\
\midrule
\multirow{3}{*}{Online CReg}
& LinUCB
& $1885.934\pm33.885$
& $\mathbf{1573.142\pm38.906}^{\ast\ast\ast}$ \\
& NeuralUCB
& $2059.076\pm29.659$
& $\mathbf{1669.303\pm81.538}^{\ast\ast\ast}$ \\
& NeuralTS
& $2074.051\pm15.427$
& $\mathbf{1705.542\pm47.198}^{\ast\ast\ast}$ \\
\midrule
\multirow{3}{*}{Cold-start Regret ($t=10$)}
& LinUCB
& $0.170\pm0.014$
& $\mathbf{0.080\pm0.011}^{\ast\ast\ast}$ \\
& NeuralUCB
& $0.173\pm0.015$
& $\mathbf{0.060\pm0.014}^{\ast\ast\ast}$ \\
& NeuralTS
& $0.175\pm0.008$
& $\mathbf{0.078\pm0.023}^{\ast\ast\ast}$ \\
\bottomrule
\end{tabular}
\end{table}

Removing dimension-level routing consistently increases both online cumulative regret and cold-start final-round regret across all three backbones, showing that TDGE's gains cannot be explained by feature-level semantic retrieval alone and specifically support the contribution of learned dimension-level routing.

\section{Regret Trajectories}
\label{app:regret_curves}
\begin{figure}[h]
\centering
\includegraphics[width=\textwidth]{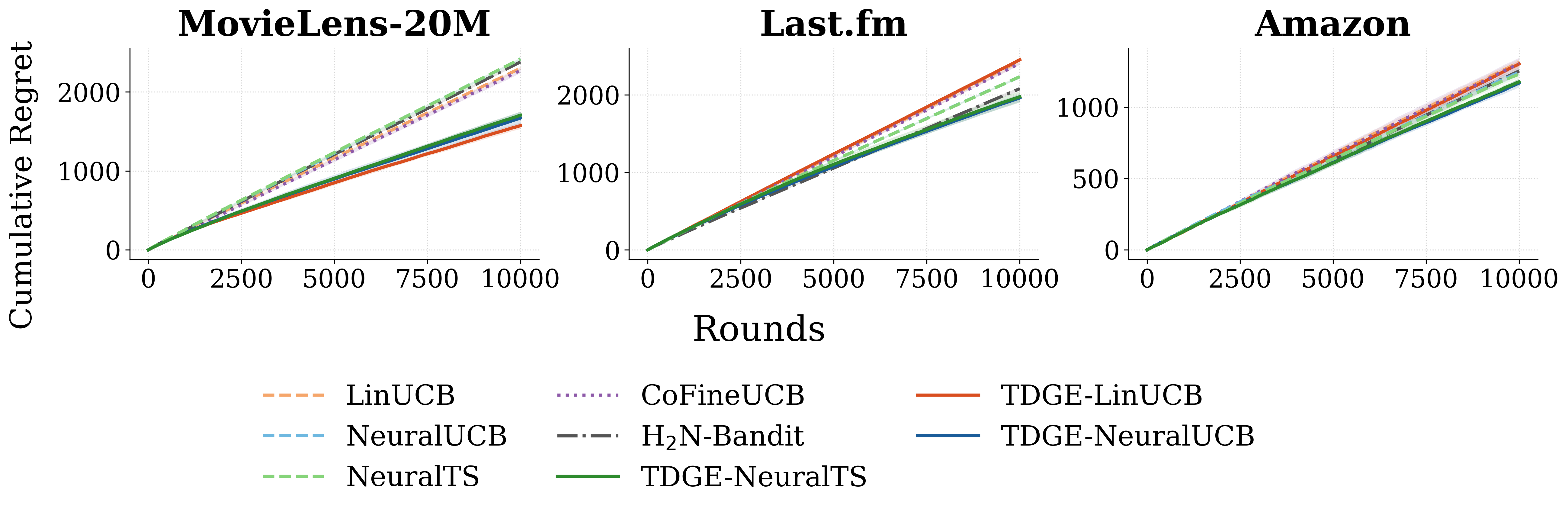}
\caption{Online cumulative regret over 10,000 interaction rounds. Curves show
the mean over five runs, and shaded regions denote one standard deviation.
Dashed curves are the item-level baselines, and solid curves are the
corresponding TDGE variants. Dotted and dash-dotted curves denote CoFineUCB
and H$_2$N-Bandit, respectively, where results are available. Lower is
better.}
\label{fig:app_online_regret_curves}
\end{figure}

Figure~\ref{fig:app_online_regret_curves} reports the complete online
cumulative-regret trajectories. TDGE reduces cumulative regret for all three
backbones on MovieLens-20M and for both neural backbones on Last.fm and
Amazon. TDGE-LinUCB remains comparable to LinUCB on Amazon but does not
improve over it on Last.fm, consistent with the endpoint comparisons in
Table~\ref{tab:main_results}. On MovieLens-20M, the corresponding TDGE
variants also maintain lower regret than CoFineUCB and H$_2$N-Bandit
throughout online training.

\begin{figure}[h]
\centering
\includegraphics[width=\textwidth]{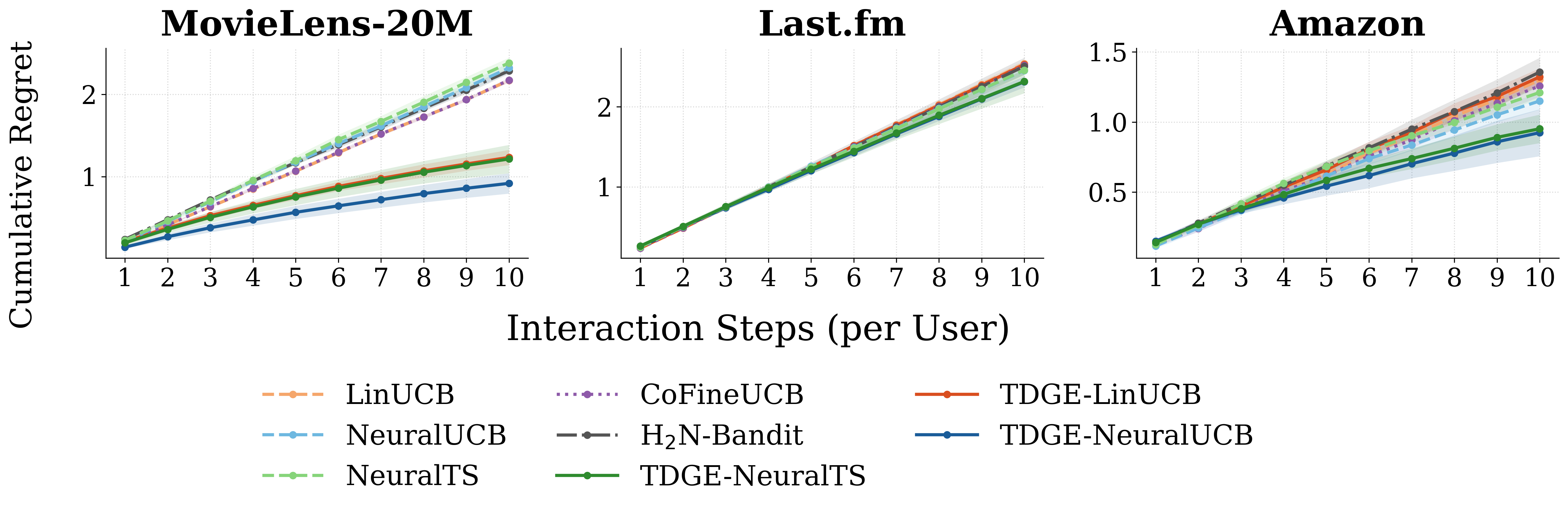}
\caption{Cold-start cumulative regret over 10 interactions per held-out user.
Within each run, regret at each step is averaged over 100 users before
accumulation. Curves show the mean over five runs, and shaded regions denote
one standard deviation. Dashed curves are the item-level baselines, and solid
curves are the corresponding TDGE variants. Dotted and dash-dotted curves
denote CoFineUCB and H$_2$N-Bandit, respectively, where results are available.
Lower is better.}
\label{fig:app_cold_start_regret_curves}
\end{figure}

Figure~\ref{fig:app_cold_start_regret_curves} shows cumulative regret over the 10-step zero-history cold-start trajectory. TDGE accumulates substantially
less regret for all three backbones on MovieLens-20M. On Last.fm, all TDGE
variants also end with numerically lower cumulative regret. On Amazon, the neural TDGE variants reduce cumulative regret, while TDGE-LinUCB is comparable to LinUCB, likely due to the limited expressiveness of the linear
reward model for capturing complex user–item interactions, as we mentioned in previous Section~\ref{sec:recommendation_results}. For completeness, we also report the regret trajectories of CoFineUCB and H$_2$N-Bandit, further supporting the structured-baseline results reported in Section~\ref{sec:structured_baselines}.
The main text reports regret at the tenth interaction to measure recommendation
quality at the end of adaptation; the cumulative trajectories additionally
show the regret incurred throughout the adaptation process.

\begin{figure}[h]
\centering
\includegraphics[width=\textwidth]{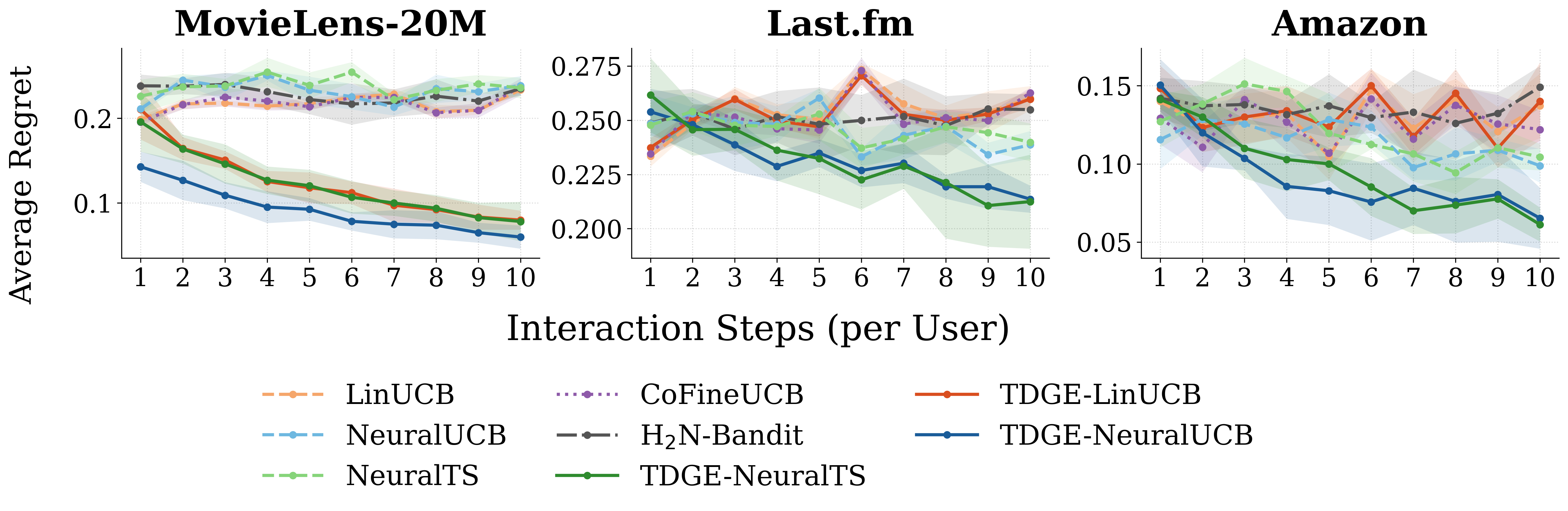}
\caption{Average instantaneous regret at each step of the 10-step cold-start
trajectory. At each step, regret is averaged over 100 held-out users within
each run. Curves show the mean over five runs, and shaded regions denote one
standard deviation. The values at step 10 correspond to the final-round
regret reported in Table~\ref{tab:main_results}. Lower is better.}
\label{fig:app_cold_start_average_regret}
\end{figure}

Figure~\ref{fig:app_cold_start_average_regret} provides the corresponding
per-step view of cold-start adaptation. Each point is the average
instantaneous regret at that interaction step, and the point at step 10 is
exactly the final-round regret reported in Table~\ref{tab:main_results}.
The trajectories verify that the final-round differences appear directly in
instantaneous regret rather than being induced by cumulative aggregation.
The reduction is most pronounced across all three backbones on MovieLens-20M
and for the neural backbones on Amazon; the Last.fm endpoints also favor the
corresponding TDGE variants.

\section{Inference-Time Analysis}
\label{app:inference_time}
TDGE introduces dimension- and feature-level scoring before item selection.
To quantify the resulting overhead at a large arm-set size, we benchmark
neural inference on an NVIDIA GeForce RTX 4090 using 101-dimensional
synthetic contexts. Both the flat backbone and its TDGE variant score 100,000
item arms per decision; TDGE additionally scores 5,000 dimension arms and
5,000 feature arms. We use 50 warm-up passes followed by 150 timed passes and
report median and 95th-percentile latency. Context tensors are constructed
before timing and retained on the GPU, so the benchmark measures model
scoring and action selection rather than data loading.

\begin{table}[htbp]
\centering
\caption{Per-decision neural inference latency with 100,000 item arms.
TDGE additionally scores 5,000 dimension arms and 5,000 feature arms.
Measurements use an NVIDIA GeForce RTX 4090 after 50 warm-up passes; 150
passes are timed.}
\label{tab:app_inference_time}
\begin{tabular}{lrr}
\toprule
Method & Median (ms) & P95 (ms) \\
\midrule
NeuralUCB      & 1.6716 & 1.7666 \\
TDGE-NeuralUCB & 1.9353 & 1.9538 \\
NeuralTS       & 1.7218 & 1.7504 \\
TDGE-NeuralTS  & 2.0404 & 2.0606 \\
\bottomrule
\end{tabular}
\end{table}

Despite the two additional routing stages, TDGE adds only 0.2637 ms for
NeuralUCB and 0.3186 ms for NeuralTS at the median, while the complete
scored decision in this benchmark remains close to 2 ms. This benchmark is
conservative with respect to TDGE's intended deployment: it assigns the same
100,000 item arms to the item-level model for both methods and therefore does
not credit TDGE for the computation saved by narrowing the item pool before
final scoring. Moreover, the dimension and feature arm sets are determined by
the metadata vocabulary and do not scale directly with the catalog size.
These results show that TDGE introduces modest scoring overhead even for a
large item set and support the practical scalability of its semantic routing
in large recommendation systems.

\clearpage
\section*{NeurIPS Paper Checklist}

\begin{enumerate}

\item {\bf Claims}
    \item[] Question: Do the main claims made in the abstract and introduction accurately reflect the paper's contributions and scope?
    \item[] Answer: \answerYes{} 
    \item[] Justification: The abstract and introduction describe TDGE as a human-inspired, model-agnostic framework that performs dimension--feature--item exploration and improves online training and cold-start adaptation. Section~\ref{sec:method} specifies the framework; Tables~\ref{tab:main_results} and~\ref{tab:structured_baselines}, Section~\ref{sec:exp_ablation}, and Appendix~\ref{app:inference_time} support the performance, robustness, and efficiency claims; and Section~\ref{sec: human-like exploration} reports dimension-persistent trajectories while explicitly avoiding claims of equivalence to human cognitive mechanisms.

\item {\bf Limitations}
    \item[] Question: Does the paper discuss the limitations of the work performed by the authors?
    \item[] Answer:  \answerYes{} 
    \item[] Justification: Section~\ref{sec: conclusion and limitation} discusses three limitations: the fixed retention sizes $K_1$ and $K_2$, the possibility that the observed effect of hierarchy depth depends on dataset structure, and the current restriction of the evaluation to contextual-bandit backbones rather than general reinforcement learning.

\item {\bf Theory assumptions and proofs}
    \item[] Question: For each theoretical result, does the paper provide the full set of assumptions and a complete (and correct) proof?
    \item[] Answer: \answerNA{} 
    \item[] Justification: The paper does not claim a formal regret bound or convergence theorem. Section~\ref{sec:method} and Appendix~\ref{app:algorithm} use mathematical expressions to define TDGE's selection, scoring, and update rules rather than to state theoretical results requiring proof.

    \item {\bf Experimental result reproducibility}
    \item[] Question: Does the paper fully disclose all the information needed to reproduce the main experimental results of the paper to the extent that it affects the main claims and/or conclusions of the paper (regardless of whether the code and data are provided or not)?
    \item[] Answer: \answerYes{} 
    \item[] Justification: Section~\ref{sec:method} and Appendix~\ref{app:algorithm} provide the selection and update equations and full pseudocode in Algorithm~\ref{alg:app_full_tdge}. Appendix~\ref{app:datasets} documents preprocessing and the offline evaluation environment, while Appendix~\ref{app:hyperparams} reports the random seeds, grid-search spaces, selected retention budgets, model configurations, and selected backbone parameters in Tables~\ref{tab:app_shared_hyper}--\ref{tab:app_tdge_agent_hyper}.

\item {\bf Open access to data and code}
    \item[] Question: Does the paper provide open access to the data and code, with sufficient instructions to faithfully reproduce the main experimental results, as described in supplemental material?
    \item[] Answer: \answerYes{} 
    \item[] Justification: MovieLens-20M, Last.fm-2K, and Amazon Reviews are publicly accessible datasets, and Appendix~\ref{app:datasets} documents their preprocessing. The repository at \url{https://github.com/LordSlime/Human-inspired-Task-Dimension-Guided-Exploration-for-Efficient-Learning} provides preprocessing, training, metric-computation, and plotting code together with reproduction instructions.

\item {\bf Experimental setting/details}
    \item[] Question: Does the paper specify all the training and test details (e.g., data splits, hyperparameters, how they were chosen, type of optimizer) necessary to understand the results?
    \item[] Answer: \answerYes{} 
    \item[] Justification: Section~\ref{sec:experiments} specifies the datasets, baselines, evaluation metrics, and paired-comparison protocol. Appendices~\ref{app:datasets} and~\ref{app:hyperparams} give the user split, zero-history cold-start construction, interaction horizons, candidate budgets, representations, network and optimizer settings, random seeds, grid-search spaces, and selected hyperparameters.

\item {\bf Experiment statistical significance}
    \item[] Question: Does the paper report error bars suitably and correctly defined or other appropriate information about the statistical significance of the experiments?
    \item[] Answer: \answerYes{} 
    \item[] Justification: Tables~\ref{tab:main_results} and~\ref{tab:structured_baselines} report mean $\pm$ sample standard deviation over five independent seeds, and significance is assessed on run-level final metrics using two-sided Welch $t$-tests. Appendix~\ref{app:regret_curves} additionally plots mean regret trajectories with one-standard-deviation bands.

\item {\bf Experiments compute resources}
    \item[] Question: For each experiment, does the paper provide sufficient information on the computer resources (type of compute workers, memory, time of execution) needed to reproduce the experiments?
    \item[] Answer: \answerYes{} 
    \item[] Justification:  Appendix~\ref{app:hyperparams} reports the CPU, GPU, system memory, and GPU memory used for the experiments, together with separate per-run wall-clock training times for the reported experiments. Appendix~\ref{app:inference_time} separately reports the hardware and measurement protocol used for the inference-time benchmark.
    
\item {\bf Code of ethics}
    \item[] Question: Does the research conducted in the paper conform, in every respect, with the NeurIPS Code of Ethics \url{https://neurips.cc/public/EthicsGuidelines}?
    \item[] Answer: \answerYes{} 
    \item[] Justification: The experiments use established public recommendation datasets described in Appendix~\ref{app:datasets} and do not collect new human-subject data. Section~\ref{sec:inspiration} uses a separately published behavioral study only as scientific motivation; no participant-level data from that study are used or analyzed in this work.

\item {\bf Broader impacts}
    \item[] Question: Does the paper discuss both potential positive societal impacts and negative societal impacts of the work performed?
    \item[] Answer: \answerYes{} 
    \item[] Justification: TDGE may improve sample efficiency and reduce low-quality recommendations during new-user adaptation (Section~\ref{sec:recommendation_results}); its explicit dimension-level routing may also aid interpretation. As with other personalized recommenders, deployment could inherit bias from interaction data and metadata or reinforce narrow content exposure, so fairness, diversity, and privacy should be evaluated in the intended application.
    
\item {\bf Safeguards}
    \item[] Question: Does the paper describe safeguards that have been put in place for responsible release of data or models that have a high risk for misuse (e.g., pre-trained language models, image generators, or scraped datasets)?
    \item[] Answer: \answerNA{} 
    \item[] Justification: The paper does not release a pretrained generative model, a newly scraped dataset, or another asset with elevated misuse potential. It releases algorithmic code and uses an existing Sentence-BERT/MPNet encoder only to embed item-descriptive features, as documented in Appendices~\ref{app:datasets} and~\ref{app:clustering}.

\item {\bf Licenses for existing assets}
    \item[] Question: Are the creators or original owners of assets (e.g., code, data, models), used in the paper, properly credited and are the license and terms of use explicitly mentioned and properly respected?
    \item[] Answer: \answerYes{} 
    \item[] Justification: The paper and supplementary materials identify
    the sources and applicable usage terms of the datasets and pretrained
    encoder. All baseline methods are cited. CoFineUCB is independently
    implemented from its published description, while H$_2$N-Bandit is
    adapted from the implementation publicly released by its original
    authors, whose source is explicitly provided. Raw third-party datasets
    are not redistributed.

\item {\bf New assets}
    \item[] Question: Are new assets introduced in the paper well documented and is the documentation provided alongside the assets?
    \item[] Answer: \answerYes{} 
    \item[] Justification: The submission releases the TDGE implementation as a new code asset through the anonymous repository linked in the abstract. Its README documents the environment, dataset preparation, training commands, evaluation scripts, result aggregation, and figure generation; no new dataset or pretrained model is released.

\item {\bf Crowdsourcing and research with human subjects}
    \item[] Question: For crowdsourcing experiments and research with human subjects, does the paper include the full text of instructions given to participants and screenshots, if applicable, as well as details about compensation (if any)? 
    \item[] Answer: \answerNA{} 
    \item[] Justification: This paper does not conduct crowdsourcing or new human-subjects research and does not use participant-level data from the behavioral study cited for motivation in Section~\ref{sec:inspiration}. All experiments reported in this paper are computational evaluations on existing recommendation datasets.


\item {\bf Institutional review board (IRB) approvals or equivalent for research with human subjects}
    \item[] Question: Does the paper describe potential risks incurred by study participants, whether such risks were disclosed to the subjects, and whether Institutional Review Board (IRB) approvals (or an equivalent approval/review based on the requirements of your country or institution) were obtained?
    \item[] Answer: \answerNA{} 
    \item[] Justification: This paper does not conduct human-subjects research or analyze participant-level data. The behavioral study cited in Section~\ref{sec:inspiration} is prior work used only to motivate the algorithm and is outside the experimental scope of this paper.

\item {\bf Declaration of LLM usage}
    \item[] Question: Does the paper describe the usage of LLMs if it is an important, original, or non-standard component of the core methods in this research? Note that if the LLM is used only for writing, editing, or formatting purposes and does \emph{not} impact the core methodology, scientific rigor, or originality of the research, declaration is not required.
    \item[] Answer: \answerNA{} 
    \item[] Justification: LLMs are not an important, original, or non-standard component of TDGE. The paper uses the off-the-shelf all-mpnet-base-v2 sentence encoder to embed feature strings during preprocessing (Appendices~\ref{app:datasets} and~\ref{app:clustering}), rather than using an LLM as part of the proposed learning or decision-making method.

\end{enumerate}

\end{document}